\documentclass[10pt,journal,compsoc]{IEEEtran}
\ifCLASSOPTIONcompsoc
\usepackage[nocompress]{cite}
\else
\usepackage{cite}
\fi
\ifCLASSINFOpdf
\else
\fi
\usepackage{amsmath,amssymb,amsthm,bm,bbm}
\usepackage{color,xcolor,colortbl}
\usepackage{graphicx,float,caption,subfig,hhline}
\usepackage{cleveref}
\usepackage{algorithm,algorithmic,multirow,MnSymbol}
\usepackage{setspace}
\usepackage{etoolbox}
\AtBeginEnvironment{algorithmic}{\onehalfspacing}

\allowdisplaybreaks[4]

\newtheorem{thm}{Theorem}

\newtheorem{lem}{Lemma}
\DeclareMathOperator{\diag}{diag}

\begin{document}
%
\title{On--the--Fly Joint Feature Selection and Classification}
%
%
%
%

\author{Yasitha~Warahena Liyanage,~\IEEEmembership{Student member,~IEEE,}
	Daphney--Stavroula~Zois,~\IEEEmembership{Member,~IEEE,}
	and~Charalampos~Chelmis,~\IEEEmembership{Member,~IEEE}
	\IEEEcompsocitemizethanks{\IEEEcompsocthanksitem Y. Warahena Liyanage and D.--S. Zois are with the Department
		of Electrical and Computer Engineering, University at Albany, SUNY,
		NY, 12222.\protect\\
		E-mail: yliyanage@albany.edu, dzois@albany.edu
		\IEEEcompsocthanksitem C. Chelmis is with the Department of Computer Science, University at Albany, SUNY,
		NY, 12222.\protect\\
		E-mail: cchelmis@albany.edu}
}

\IEEEtitleabstractindextext{%
	\begin{abstract}
		Joint feature selection and classification in an online setting is essential for time--sensitive decision making. However, most existing methods treat this coupled problem independently. Specifically, online feature selection methods can handle either streaming features or data instances offline to produce a fixed set of features for classification, while online classification methods  classify incoming instances using full knowledge about the feature space. Nevertheless, all existing methods  utilize a set of features, common for all data instances, for classification. Instead, we propose a framework to perform joint feature selection and classification on--the--fly, so as to  minimize the number of features evaluated for every data instance and maximize classification accuracy.  We derive the optimum solution of the associated optimization problem and analyze its structure. Two algorithms are proposed, ETANA and F--ETANA, which are based on the optimum solution and its properties. We evaluate the performance of the proposed algorithms on several public datasets, demonstrating (i) the dominance of the proposed algorithms over the state--of--the--art, and (ii) its applicability to broad range of application domains including clinical research and natural language processing.  		
	\end{abstract}
	
	}

\maketitle

\IEEEdisplaynontitleabstractindextext

%
\IEEEpeerreviewmaketitle

\IEEEraisesectionheading{\section{Introduction}\label{sec:introduction}}


%
%
%
%
\IEEEPARstart{F}{eature} selection is the process of selecting a subset of the most informative features from a large set of potentially redundant features with the objective of maximizing classification accuracy, alleviating the effect of the curse of dimensionality, speeding up the training process and improving interpretability~\cite{guyon2003introduction,liu2005toward}. 

\begin{figure}[!tb]
	\centering
	\includegraphics[width=\columnwidth]{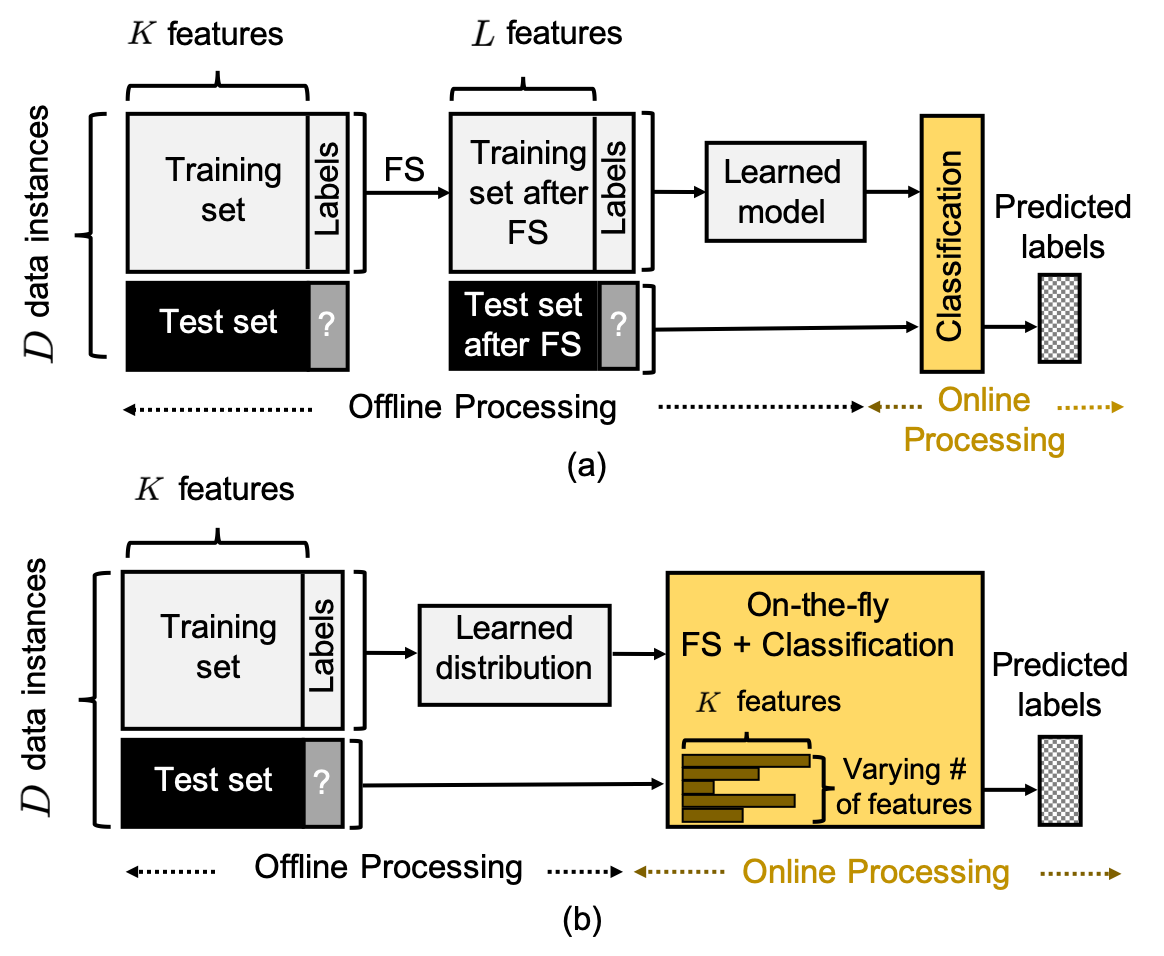} 
	\vspace{-0.2in}
	\caption{Using a $K \times D$ matrix,  (a) existing  feature selection (FS) methods extract a set $L << K$ of features, which is common for all data instances during classification. In contrast, (b) the proposed on--the--fly joint feature selection and classification approach utilizes a varying number of features to classify each data instance online using a model learned offline from all features.}
	\label{fig:fs_classify}
\end{figure}

Most existing work on feature selection~\cite{perkins2003online,zhou2005streaming,wu2012online,yu2014towards,zhou2019ofs,eskandari2016online,zhou2019online,javidi2019online,hoi2012online,wang2014onine,wu2017large,li2018feature,wang2017feature,zhang2018muse,chandrashekar2014survey,saeys2007review,hu2018survey,hu2018survey,alnuaimi2019streaming} extracts
a subset of discriminative features that can  globally describe the data well, where the same feature subset is used to classify all instances during classification (see Fig.~\ref{fig:fs_classify}(a)). Only a handful of feature selection methods have considered the feature evaluation cost  and costs associated with misclassification, which play a key role in many real--world applications \cite{bolon2014framework,shu2016multi}.  On the other hand, existing online classification techniques~\cite{rosenblatt1958perceptron,novikoff1963convergence,crammer2006online,duchi2011adaptive,van2016metagrad,luo2016efficient,zhang2018adaptive,li2002relaxed,wang2013cost,zhao2018adaptive,hoi2018online} update model parameters by examining incoming data instances one at a time; such methods are not only affected by noisy and missing data, but also face scalability constraints~\cite{hoi2018online}. 

In a depart from existing feature selection and classification methods, we study the problem of on--the--fly joint feature selection and classification (OJFC)  in an online setting. Specifically, the goal of OJFC is to minimize the number of feature evaluations for classification individually for each instance, while achieving high classification accuracy across all instances. This is particularly important and necessary for real--world applications requiring time--sensitive decisions such as weather forecasting \cite{zhang2016short}, transportation \cite{yang2013feature}, stock markets prediction \cite{lee2009using}, clinical research \cite{ding2005minimum}, and natural disasters prediction \cite{seo2014feature}. Therefore, the proposed method  utilizes a varying number of features to classify each data instance (see Fig.~\ref{fig:fs_classify}(b)).

To address the challenges associated with the problem of OJFC, we define an optimization problem which simultaneously minimizes the number of features evaluated and maximizes classification accuracy.  The solution to this optimization problem leads to an approach that sequentially reviews features and classifies a data instance once it determines that including additional features cannot further improve the quality of classification. However, the computational complexity of the optimum solution increases exponentially with the number of classes in multi--class classification tasks.  
To improve the scalability of our approach, we propose an efficient implementation, which exploits the structure of the optimum solution. Specifically, the functions related
to the optimum solution are shown to be concave, continuous and piecewise linear on the domain of a sufficient statistic.
As a result, the optimum solution exhibits a threshold structure to decide between continuing the feature evaluation process and stopping. A stochastic gradient algorithm is utilized to estimate the optimal linear thresholds.  Extensive experimental evaluation using seven publicly available datasets shows the superiority of the proposed approach in terms of classification accuracy, average number of features used per data instance, and time required for joint feature selection and classification compared to the state--of--the--art. Further, our evaluation results indicate that the proposed efficient implementation drastically reduces training time without a drop in accuracy as compared to the optimum solution. All proofs are included in Appendices A and B. 


\section{Related Work}
In this section, we summarize the most relevant prior work on (i) online feature selection and (ii) online classification techniques. 
\subsection{Online feature selection}
In contrast to offline feature selection methods that are designed for static datasets with fixed number of features and data samples, online feature selection methods are capable of handling either streaming features or streaming data samples to choose a subset of features from a larger set of potentially redundant features~\cite{li2018feature,hu2018survey,alnuaimi2019streaming}.  Online feature selection methods can be generally grouped into two groups: \\
\textbf{(a) Streaming Features}: In  this branch of online feature selection problems, the number of data instances is considered constant while features arrive one at a time~\cite{perkins2003online,zhou2005streaming,wu2012online,yu2014towards,eskandari2016online,zhou2019ofs,javidi2019online,zhou2019online}. In~\cite{perkins2003online}, a newly arriving feature is selected if the improvement in the model is greater than a predefined threshold. \cite{wu2012online,yu2014towards} try to extract features in the  Markov blanket of the class variable  using a forward algorithm, where thresholds on probability approximations to measure conditional independence (e.g. G$^2$--test~\cite{wu2012online}, Fisher's Z--test\cite{yu2014towards}) are employed. Such threshold--based methods~\cite{perkins2003online,zhou2005streaming,wu2012online,yu2014towards} require prior information about the feature space~\cite{zhou2019ofs}. 
Recently, rough set theory based methods~\cite{eskandari2016online,zhou2019ofs,javidi2019online,zhou2019online} have been explored.  Such methods do not require any domain knowledge~\cite{eskandari2016online}. However, methods proposed in~\cite{eskandari2016online,javidi2019online} are not applicable to numerical features, while methods~\cite{zhou2019ofs,zhou2019online} are much slower in feature selection compared to the state--of--the--art streaming feature selection methods. \\
\textbf{(b) Streaming Data}: In this problem setting, the number of features is considered constant, while data instances arrive over time~\cite{hoi2012online,wang2014onine,wu2017large}. Such methods~\cite{hoi2012online,wang2014onine,wu2017large}, are limited to binary classification and/or impose hard constraints on the number of non--zero elements in the model, requiring the user to define the number of features that need to be selected \textit{a priori}~\cite{hoi2018online}. 
\subsection{Online Classification}
Online classification methods, also referred to as online learning, use sequentially arriving data to update the function of a classifier. This is in contrast to batch learning techniques where a  collection of training data is used to train a classifier offline, without further updates once training is complete~\cite{hoi2018online}. Most widely used online learning methods~\cite{rosenblatt1958perceptron,novikoff1963convergence,crammer2006online} are either limited to binary classification~\cite{rosenblatt1958perceptron,novikoff1963convergence} or require solving a complex optimization problem at each iteration, and require prior information to tune parameters in the model~\cite{crammer2006online}. On the other hand, traditional gradient based methods~\cite{duchi2011adaptive,van2016metagrad,luo2016efficient,zhang2018adaptive} not only require to compute the gradient of a cost function, but also require to solve an optimization problem at each iteration. Cost--sensitive extensions of traditional online classification methods, which account for misclassification costs have been recently explored~\cite{li2002relaxed,crammer2006online,wang2013cost,zhao2018adaptive}. Unlike our approach, ~\cite{li2002relaxed,crammer2006online} do not optimize the misclassification cost directly~\cite{hoi2018online}, while \cite{li2002relaxed,wang2013cost,zhao2018adaptive} are limited to binary classification. Last but not least, most existing methods are highly susceptible to noise and/or incomplete data~\cite{hoi2018online}. 
\section{On--the--Fly Joint Feature Selection and Classification} \label{sec:PF}

Consider a set $\mathcal{S}$ of data instances, with each data instance $s \in \mathcal{S} $ being described using an assignment of values $f = \{f_1, f_2, \dots, f_K \}$ to a set $F = \{ F_1,F_2,\dots,F_K\}$ of $K$ features. Each data instance $s$ is drawn from some probability distribution over the feature space such that for each assignment $f$ to $F$, we have a probability $P(F=f)$. Further, each instance $s$ may belong to one of $N$ classes, with corresponding \textit{a priori} probability $P(T =T_i )  = p_i$ for each assignment $T_i, i = 1, 2, \dotsc, N$, of the class variable $T$. Moreover, coefficients $c_k >0, k = 1, 2, \dotsc, K$, represent the cost of evaluating features $F_k$, respectively, and coefficients $M_{i,j} \geqslant 0, i,j\in \{1,\dots,N\}$, denote the misclassification cost of selecting class $T_j$ when class $T_i$ is true. 

To select one out of $N$ possible classes for each data instance $s$, our proposed approach evaluates features sequentially, where at each step it has to decide between stopping and continuing the feature evaluation process based on the accumulated information thus far and the cost of evaluating the remaining features. Herein, we introduce a pair of random variables $(R,D_R)$, where $0 \leq R \leq K$ (referred to as \textit{stopping time} \cite{shiryaev2007optimal} in decision theory) denotes the feature at which the framework assigns $s$ to a specific class, and $D_R \in \{1,\dots,N\}$, which depends on $R$, denotes the possibility to select among the $N$ classes. The event $\{R=k\}$ depends only on the feature set $\{F_1, F_2, \dots, F_k \} $, whereas the event $\{D_R=j\}$ represents choosing class $T_j$ based on information accumulated up to feature $R$. The goal is to select random variables $R$ and $D_R$ by solving the following optimization problem:
\begin{equation}\label{eq:optimization_problem}
\underset{R,D_R}{\text{minimize}} \ J(R,D_R),
\end{equation}
where the cost function is defined as:
\begin{equation}\label{eq:optimization_function}
J(R,D_{R}) \triangleq \mathbb{E} \bigg \lbrace \sum_{k = 1}^{R} c_{k}  \bigg \rbrace + \sum_{j = 1}^{N} \sum_{i = 1}^{N} M_{ij} P(D_{R} = j,T=T_{i}),
\end{equation}
in which the first term denotes the cost of evaluating features, and the second term penalizes misclassification errors. 
%

To solve the optimization problem defined in Eq. (\ref{eq:optimization_problem}), we define a sufficient statistic of accumulated information, the \textit{a posteriori probability} vector $\pi_k$, as follows:
\begin{align}
\pi_{k} \triangleq [\pi_{k}^1,\pi_{k}^2,\dots,\pi_{k}^N]^T,
\end{align}
where the $k$th feature is evaluated to generate outcome $f_{k}$, and $\pi_{k}^i \triangleq P(T_i | F_{1}, \dotsc, F_{k})$. To simplify the notation, $P(T_i | F_{1}, \dotsc, F_{k})$ is used in lieu of $P(T=T_i | F_{1}=f_1, \dotsc, F_{k}=f_k)$ subsequently.  Assuming that features in set $F$ are independent given the class variable $T$\footnote{Even though validation of this assumption is beyond the scope of this paper, we find our proposed method to work well in practice.}, $\pi_{k}$ can be computed recursively as in Lemma \ref{lem:posterior_probability}.
\begin{lem}\label{lem:posterior_probability}
	The a posteriori probability vector $\pi_{k} \in [0,1]^N$ can be recursively computed as:
	\begin{equation}\label{eq:update_rule}
	\pi_{k} = \frac{\diag(\Delta_k(F_k)) \pi_{k-1}}{\Delta_k^T(F_k) \pi_{k-1}},
	\end{equation}
	where $\Delta_k(F_k) \triangleq [P(F_{k}| T_1),P(F_{k}| T_2),\dots,P(F_{k}| T_N)]^T$, $\diag(A)$ denotes a diagonal matrix with diagonal elements being the elements in vector $A$, and $\pi_{0} \triangleq [p_1,p_2,\dots,p_N]^T$.
\end{lem}
\noindent Next, we simplify the probability $P(D_{R} = j,T=T_{i})$ exploiting the definition of the \textit{a posteriori} probability $\pi_R^i$.  
\begin{lem}\label{lem:indicator}
	Based on the fact that $x_{R} = \sum_{k = 0}^{K} x_{k} \mathbbm{1}_{\lbrace R = k\rbrace}$ for any sequence of random variables $\lbrace x_{k} \rbrace$, where $\mathbbm{1}_{A}$ is the indicator function for event $A$ (i.e., $\mathbbm{1}_{A} = 1$ when $A$ occurs, and $\mathbbm{1}_{A} = 0$ otherwise), the probability $P(D_{R} = j,T_{i})$ can be written as follows:
	\begin{equation}
	P(D_{R} = j,T_{i}) = \mathbb{E} \left \lbrace \pi_{R}^i  \mathbbm{1}_{\lbrace D_{R} = j \rbrace} \right \rbrace.
	\end{equation}
\end{lem}

\noindent Using Lemma~\ref{lem:indicator},  the average cost in Eq.~(\ref{eq:optimization_function}) can be written compactly as:
\begin{small}
	\begin{align}\label{eq:optimization_function_new}
	J(R,D_{R}) = \mathbb{E} \left \lbrace \sum_{k = 1}^{R} c_{k} + \sum_{j = 1}^{N} \Big( \sum_{i = 1}^{N} M_{ij} \pi_{R}^i \Big ) \mathbbm{1}_{\lbrace D_{R} = j \rbrace} \right \rbrace,
	\end{align}
\end{small}
\noindent which in turn can be rewritten as follows:
\begin{equation}\label{eq:cost_function_new}
J(R,D_{R}) = \mathbb{E} \left \lbrace \sum_{k = 1}^{R} c_{k} + \sum_{j = 1}^{N}  M_j^T \pi_R \mathbbm{1}_{\lbrace D_{R} = j \rbrace} \right \rbrace,
\end{equation}
where $M_j \triangleq [M_{1,j},M_{2,j},\dots,M_{N,j}]$.

To obtain the optimum stopping time $R$, we must first obtain the optimum decision rule $D_R$ for any given $R$. In the process of finding the optimum decision rule, we need to find a lower bound (independent of $D_R$)  for the second term inside the expectation in Eq. (\ref{eq:cost_function_new}), which is the part of the equation that depends on $D_R$.
Theorem~\ref{thm:optimal_selection_strategy} provides such bound.

\begin{thm}\label{thm:optimal_selection_strategy}
	For any classification rule $D_{R}$ given stopping time $R$, $\sum_{j = 1}^{N}  M_j^T \pi_R \mathbbm{1}_{\lbrace D_{R} = j \rbrace} \geqslant g(\pi_{R})$, where $g(\pi_{R}) \triangleq \min_{1 \leqslant j \leqslant N} \big [  M_j^T \pi_R  \big ]$. The optimum rule is defined as follows:
	\begin{equation}\label{eq:optimal_classification_strategy}
	D_{R}^{optimum} = {\arg \min}_{1 \leqslant j \leqslant N} \big [ M_j^T \pi_R  \big ].
	\end{equation}
\end{thm}

\noindent From Theorem~\ref{thm:optimal_selection_strategy}, we conclude that:
\begin{align}
J(R,D_{R}) &\geqslant J(R,D_{R}^{optimum }), \textnormal{ where}\nonumber  \\
J(R,D_{R}^{optimum })  &= \min_{D_{R}} J(R, D_{R}).
\end{align}
Thus, we can reduce the cost function in Eq. (\ref{eq:cost_function_new}) to one which depends only on the stopping time $R$ as follows:
\begin{equation}
\label{Eq.reduced_cost}
\widetilde{J}(R) = \mathbb{E} \left \{ \sum_{k = 1}^{R} c_{k} + g(\pi_{R}) \right \}.
\end{equation}
To optimize the cost function in Eq.~(\ref{Eq.reduced_cost}) with respect to $R$, we need to solve the following optimization problem:
\begin{equation}
\min_{R\geqslant 0} \widetilde{J}(R)  = \min_{R\geqslant 0} \mathbb{E} \left \{ \sum_{k = 1}^{R} c_{k} + g(\pi_{R}) \right \}.
\end{equation}
Since $R \in \lbrace 0, 1, \dotsc, K \rbrace$, the optimum strategy consists of a maximum of $K + 1$ stages, where the optimum solution must minimize the corresponding average cost going from stages $0$ to $K$. The solution can be obtained using \textit{dynamic programming}~\cite{BertsekasDPOC05}.

\begin{thm}\label{thm:dp_programming}
	For $k = K-1, \dotsc, 0$, function $\bar{J}_{k}(\pi_{k})$  is related to $\bar{J}_{k+1}(\pi_{k+1})$ through the equation:
	\begin{align}\label{eq:dp_programming}
	\bar{J}_{k}(\pi_{k}) = \min \bigg [& g(\pi_{k}), c_{k+1} + \sum_{F_{k+1}} \Delta_{k+1}^T(F_{k+1}) \pi_{k}  \nonumber \\
	&\times \bar{J}_{k+1} \bigg (  \frac{\diag \big(\Delta_{k+1}(F_{k+1}) \big) \pi_{k} }{\Delta_{k+1}^T(F_{k+1}) \pi_{k} }\bigg )  \bigg ],
	\end{align}
	\noindent
	where $\bar{J}_{K}(\pi_{K}) = g(\pi_{K})$.
\end{thm}

The optimum stopping strategy derived from Eq.~(\ref{eq:dp_programming}) has a very intuitive structure. Specifically, it stops at  stage $k$, where the cost of stopping (the first expression in the minimization) is no greater than the expected cost of continuing given all information accumulated at the current stage $k$ (the second expression in the minimization). Equivalently, at each stage $k$, our method faces two options given $\pi_{k}$: (i) stop evaluating features and select optimally between the $N$ classes, or (ii) continue with the next feature. The cost of stopping is $g(\pi_{k})$, whereas the cost of continuing is $c_{k+1} + \sum_{F_{k+1}} \Delta_{k+1}^T(F_{k+1}) \pi_{k}\times 
\bar{J}_{k+1} \bigg (  \frac{\diag \big(\Delta_{k+1}(F_{k+1})  \big) \pi_{k}}{\Delta_{k+1}^T(F_{k+1}) \pi_{k}} \bigg )$.
%

Based on Lemma \ref{lem:posterior_probability}, and Theorems \ref{thm:optimal_selection_strategy} and \ref{thm:dp_programming},   we present \textbf{ETANA}, an on--the--fly f\underline{E}ature selec\underline{T}ion and cl\underline{A}ssificatio\underline{N} \underline{A}lgorithm. 
Initially, the posterior probability vector $\pi_{0}$ is set to $[p_1,p_2,\dots,p_N]$, and the two terms in Eq.~(\ref{eq:dp_programming}) are compared. If the first term is less than or equal to the second term, ETANA classifies the instance under examination to the appropriate class, based on the optimum rule in Eq.~(\ref{eq:optimal_classification_strategy}). Otherwise, the first feature is evaluated. ETANA repeats these steps until either it decides to classify the instance using $<K$ features, or using all $K$ features.

To implement ETANA, we first need to solve the \textit{dynamic programming} recursion in Eq.~(\ref{eq:dp_programming}). This can be achieved by quantizing the interval $[0,1]$ over $N$ values such that $\sum_{i=1}^{N} \pi_{k}^i = 1$ to generate different possible vectors $\pi_{k}$. We can then compute a $ (K+1) \times d$ matrix, where each row contains values of the function $\bar{J}_{k}(\pi_{k}), k = 0, 1, \dotsc, K$, evaluated using Theorem \ref{thm:dp_programming} for all possible $d$ vectors of $\pi_{k}$. Although this computation requires only \emph{a priori} information, the size of this matrix (i.e., $d$)  grows exponentially with the number $N$ of classes, resulting in a computationally expensive solution. To address this challenge, we propose an efficient implementation of ETANA in Section \ref{sec.altopt}.
 
 \section{Efficient Implementation of ETANA}\label{sec.altopt}
 
 \begin{figure}[t]
 	\begin{center}
 		\includegraphics[width=0.8\linewidth]{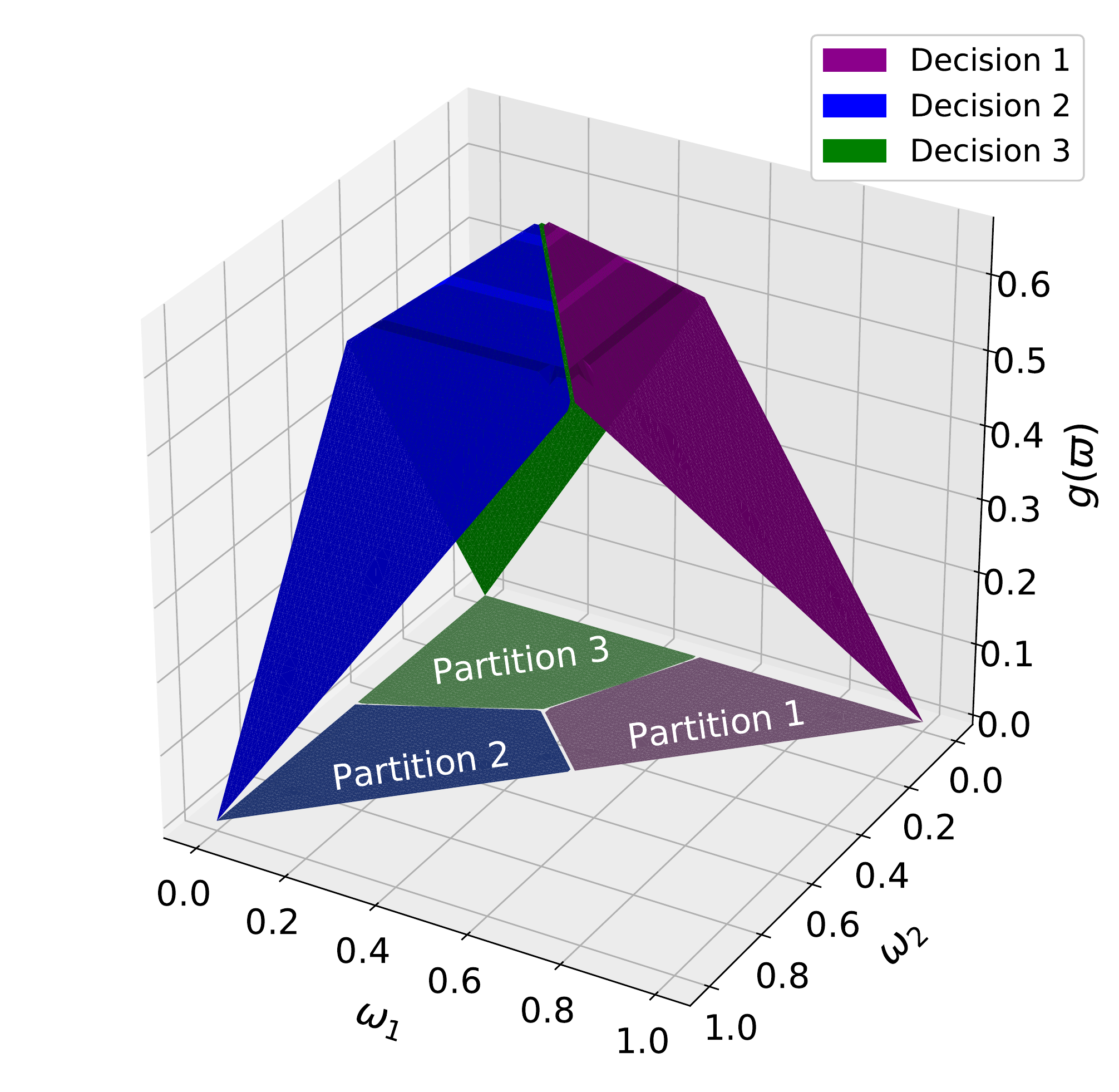} 
 		\vspace{-0.1in}
 		\caption{Illustration of Lemma~\ref{lem:g_function} when the number $N$ of classes equals to $3$, with misclassification costs $M_1 = [0,1,1], M_2 = [1,0,1]$ and $M_3 = [1,1,0]$. } 
 		\label{fig:g_w}
 	\end{center}
 	\vspace{-0.1in}
 \end{figure}
 
 \begin{figure*}[t]
 	\begin{center}
 		\includegraphics[width=0.8\textwidth]{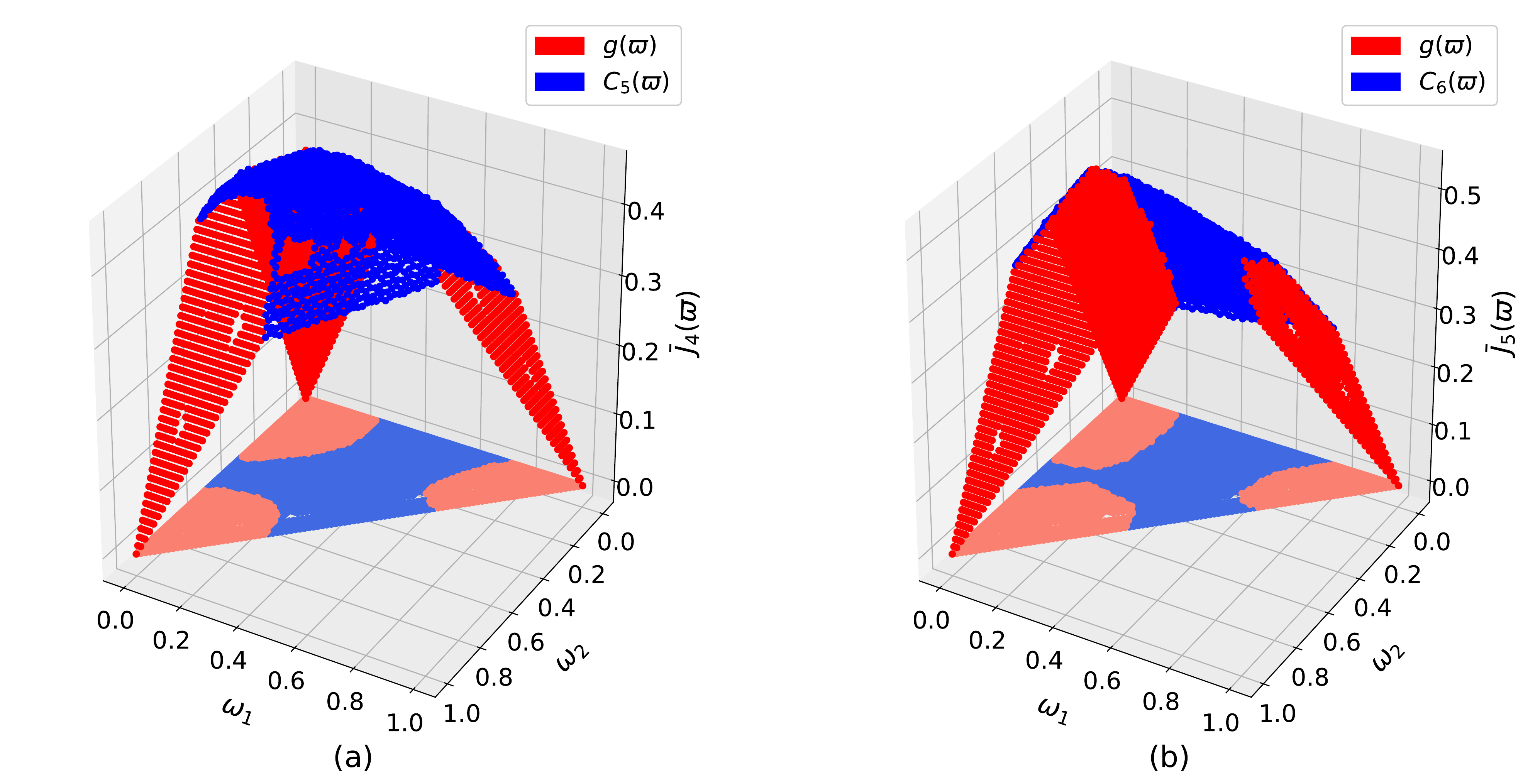} 
 		\vspace{-0.1in}
 		\caption{Illustration of Theorem~\ref{thm:alternative} (better seen in color): (a) at stage $4$ and (b) at stage $5$, using MLL dataset with misclassification costs $M_1 = [0,1,1], M_2 = [1,0,1]$ and $M_3 = [1,1,0]$. The MLL dataset contains 72 samples, each of which comprises 5848 gene expression values belonging to one of 3 diagnostic  classes~\cite{yang2006stable}.  In both cases, the blue region corresponds to continuation to the next stage, while the red region corresponds to stopping.} 
 		\label{fig:s_c}
 	\end{center}
 	\vspace{-0.1in}
 \end{figure*}
 
 In this section, we present a fast version of ETANA, namely, \textbf{F--ETANA}, that exploits structural properties of the  optimum classification rule in Eq.~(\ref{eq:optimal_classification_strategy}) and the optimum stopping strategy in Eq.~(\ref{eq:dp_programming}). 
 
 Consider a general form of the function $g(\pi_{R})$ used to derive the optimum classification rule in Eq.~(\ref{eq:optimal_classification_strategy}) as follows:
 \begin{equation}\label{eq:g_function}
 	g(\varpi)\triangleq \min_{1 \leqslant j \leqslant N} \big [   M_j^T \varpi \big ], \varpi \in [0,1]^{N},
 \end{equation}
 where $\varpi = [\omega_1,\dots,\omega_N]^T $, such that $\omega_i \geqslant 0, \sum_{i=1}^{N} \omega_i =1$. Here, the domain of $g(\varpi)$ is the probability space of $\varpi$, which is a $N-1$ dimensional unit simplex. Function $g(\varpi)$ has some interesting properties as described in Lemma~\ref{lem:g_function}. 
 
 \begin{lem}\label{lem:g_function}
 	Function $g(\varpi)$ is concave, continuous, and piecewise linear and consist of at most $N$ hyperplanes. 
 \end{lem}
 

Fig.~\ref{fig:g_w}  shows a visualization of Lemma~\ref{lem:g_function}, when $N=3$, so that the domain of $g(\varpi)$ is a $2$--dimensional unit simplex (i.e., an equilateral triangle).  Next, we consider the general form of the optimum stopping strategy in Eq.~(\ref{eq:dp_programming}) as follows:
 \begin{align}\label{eq:general_J}
 	\bar{J}_{k}(\varpi) =\min \bigg [& g(\varpi), c_{k+1} + \sum_{F_{k+1}}  \Delta_{k+1}^T(F_{k+1}) \varpi  \nonumber \\ & \times \bar{J}_{k+1} \bigg (  \frac{\diag \big(\Delta_{k+1}(F_{k+1}) \big) \varpi }{\Delta_{k+1}^T(F_{k+1}) \varpi} \bigg )  \bigg ].
 \end{align}
 Lemma~\ref{lem:j_function} summarizes the key properties enjoyed by this function.
 
 \begin{lem}\label{lem:j_function}
 	The functions $\bar{J}_{k}(\varpi), k=0,\dots,K-1$, are concave, continuous, and piecewise linear. 
 \end{lem}
 
 The fact that $g(\varpi)$ and $\bar{J}_{k}(\varpi)$ are  concave and piecewise linear allows for a compact representation of these functions. Recall that according to Theorem \ref{thm:dp_programming}, we stop at stage $k$ whenever $g(\varpi) \leq \bar{\mathcal{C}}_{k+1}(\varpi)$, where $\bar{\mathcal{C}}_{k+1}(\varpi)$ is the optimum cost--to--go at stage $k$ given by $c_{k+1} + \sum_{F_{k+1}} \Delta_{k+1}^T(F_{k+1}) \varpi \times 
 \bar{J}_{k+1} \bigg (  \frac{\diag \big(\Delta_{k+1}(F_{k+1})  \big) \varpi }{\Delta_{k+1}^T(F_{k+1}) \varpi} \bigg )$.
 In particular, to decide between continuing and stopping, it is sufficient to keep track of the thresholds at the intersections of $g(\varpi)$ with every $\bar{\mathcal{C}}_{k+1}(\varpi)$ as stated in Theorem~\ref{thm:alternative} below.
 
 \begin{thm}\label{thm:alternative}
 	At every stage $k$, there exists at most $N$ threshold curves that separate the unit simplex into regions which alternatively switch between continuation to the next stage and stopping. In particular, the region starting from every corner of the $N-1$ dimensional unit simplex always corresponds to stopping the feature evaluation process.
 \end{thm}
 
 Theorem~\ref{thm:alternative} turns out to be very important. Specifically, the region where the \textit{a posteriori probability vector} $\pi_{R}$ falls into will help decide between continuing to the next stage or stopping.
 This provides an alternative fast implementation of the optimum solution using thresholds.  Fig.~\ref{fig:s_c} shows a visualization of Theorem~\ref{thm:alternative}; both sub--figures contain maximum number of threshold curves (i.e., $3$ since $N=3$).  

 \subsection{Stochastic Gradient Algorithm for Estimating Optimum Linear Thresholds}
 
We propose a stochastic gradient algorithm to estimate the threshold curves described in Theorem~\ref{thm:alternative}. 
For ease of implementation, we restrict the approximation to linear threshold curves of the form given in Eq.~(\ref{eq:threshold}). 
 
Let  $\theta_{\widetilde{D}_{\widetilde{R}}}\triangleq  [\theta_{\widetilde{D}_{\widetilde{R}}}^1,\theta_{\widetilde{D}_{\widetilde{R}}}^2, \dots, \theta_{\widetilde{D}_{\widetilde{R}}}^{N}] $ denote the parameters of a linear hyperplane, where $\widetilde{R}$ is the number of features evaluated so far, and $\widetilde{D}_{\widetilde{R}}= j, \widetilde{R} \in \{0,1,\dots,K\}, j\in \{1,\dots N\}$ represents a decision choice. Then, the decision $Z_{\theta}$ to ``stop" or ``continue" at each stage $\widetilde{R}$ under the decision choice $\widetilde{D}_{\widetilde{R}} =j$, as function of  $\varpi$, is defined as follows:
 \begin{align}\label{eq:threshold}
 	Z_{\theta_{\widetilde{D}}} (\varpi)=  \left\{
 	\begin{array}{l l}
 		\text{stop},&\text{if }  \theta_{\widetilde{D}_{\widetilde{R}} }^T \varpi \leqslant 0\\
 		\text{continue}, &\text{otherwise} \\
 	\end{array}. \right. 
 \end{align}
 
Decision $Z_{\theta_{\widetilde{D}}}$ is indexed by $\theta_{\widetilde{D}}$ to show the explicit dependency of the parameters on the decision, where $\theta_{\widetilde{D}} \triangleq [\theta_{\widetilde{D}_0}, \theta_{\widetilde{D}_1} \dots, \theta_{\widetilde{D}_{K-1}}] \in  \mathbb{R}^{K \times N}, \widetilde{D}_{\widetilde{R}} = j, \forall \widetilde{R} $, is the concatenation of $\theta_{\widetilde{D}_{\widetilde{R}} }$ vectors, one for each stage $\widetilde{R}$. Now,  recall the cost function in Eq.~(\ref{Eq.reduced_cost}). Since we are interested  in finding linear thresholds for each decision choice $\widetilde{D}_{\widetilde{R}}  =j$ independently, we use a modified version of the cost function in Eq.~(\ref{Eq.reduced_cost}) as follows:
 \begin{equation}
 	\label{eq:modified_cost}
 	\widetilde{H}(\theta_{\widetilde{D}}) = \mathbb{E}_{Z_{\theta_{\widetilde{D}}}} \left \{ \sum_{k = 1}^{\widetilde{R}} c_{k} + M_{\widetilde{D}_{\widetilde{R}} }\pi_{\widetilde{R}} \right \}.
 \end{equation}
 
 \begin{algorithm}
 	\caption{Stochastic Gradient Algorithm for Estimating Optimal Linear Thresholds}
 	\begin{algorithmic}[1]
 		\renewcommand{\algorithmicrequire}{\textbf{Require:}}
 		\renewcommand{\algorithmicensure}{\textbf{Output:}}		
 		\REQUIRE Initial parameters $\theta_{\widetilde{D},0}$
 		\ENSURE Optimal parameters $\theta_{\widetilde{D},opt}$
 		\FOR{iterations $t=0,1,2,\dots$}	
 		\STATE Evaluate $\widetilde{H}(\theta_{\widetilde{D},t} + \beta_t\alpha_t )$ and $\widetilde{H}(\theta_{\widetilde{D},t} - \beta_t\alpha_t )$  using Function~\ref{Algo:algo2} based on Eq.~(\ref{eq:modified_cost})
 		\STATE Estimate $\hat{\nabla}_{\theta_{\widetilde{D}}} \widetilde{H}(\theta_{\widetilde{D},t})$ using Eq.~(\ref{eq:SPSA_grad})
 		\STATE Update  $\theta_{\widetilde{D},t}$ to $\theta_{\widetilde{D}, t+1}$ using Eq.~(\ref{eq:SPSA_update})
 		\STATE Stop if $|| \hat{\nabla}_{\theta_{\widetilde{D}}} \widetilde{H}(\theta_{\widetilde{D},t})||_2 \leqslant  \varepsilon$ or maximum number  $t_{\text{max}}$ of iterations reached
 		\ENDFOR 
 		\RETURN $\theta_{\widetilde{D},{opt}}$
 	\end{algorithmic}
 	\label{Algo:algo1}
 \end{algorithm}
 
 Algorithm~\ref{Algo:algo1} generates a sequence of estimates $\theta_{\widetilde{D},t}$ by computing the gradient $\nabla_{\theta_{\widetilde{D}}} \widetilde{H}(\theta)$. Here, $\theta_{\widetilde{D},t}$ denotes the estimate of $\theta_{\widetilde{D}}$ at iteration $t$. Although evaluating the gradient in closed form is intractable due to the non--linear dependency of $\widetilde{H}(\theta_{\widetilde{D}})$ and $\theta_{\widetilde{D}}$, estimate $\hat{\nabla}_{\theta_{\widetilde{D}}} \widetilde{H}(\theta_{\widetilde{D}})$ can be computed using a simulation--based gradient estimator. For simplicity, we opted for the SPSA algorithm~\cite{spall2005introduction}, among the several  simulation--based gradient estimators in the literature~\cite{pflug2012optimization}. SPSA algorithm estimates the gradient at each iteration $t$ using a finite difference method and a random direction $\alpha_t$,  as follows:
 \begin{align}\label{eq:SPSA_grad}
 \hat{\nabla}_{\theta_{\widetilde{D}}} \widetilde{H}(\theta_{\widetilde{D},t}) = \frac{\widetilde{H}(\theta_{\widetilde{D},t} + \beta_t\alpha_t )- \widetilde{H}(\theta_{\widetilde{D},t} - \beta_t\alpha_t )}{2 \beta_t} \alpha_t,
 \end{align} 
 where $\alpha_t^i =  \left\{
 \begin{array}{l l}
 -1,&\text{with probability } 0.5\\
 +1, &\text{with probability } 0.5 \\
 \end{array} \right.$.
 
 Using the gradient estimate in Eq.~(\ref{eq:SPSA_grad}), parameter $\theta_{\widetilde{D},t}$ is updated as follows:
 \begin{align}\label{eq:SPSA_update}
 	\theta_{\widetilde{D},t+1} = \theta_{\widetilde{D},t} - a_t \hat{\nabla}_{\theta_{\widetilde{D}}} \widetilde{H}(\theta_{\widetilde{D},t}),
 \end{align}
 where $a_t$ and $\beta_t$ are typically chosen as in \cite{spall2005introduction}:
 \begin{equation}
 	\begin{aligned}
 		a_t &=  \varepsilon  (t+1+\varsigma)^{-\kappa }, \qquad 0.5 < \kappa \leqslant 1, \qquad \varepsilon,\varsigma >0, \\
 		\beta_t &= \mu (t+1)^{-\upsilon },  \qquad 0.5 < \upsilon \leqslant 1, \qquad \mu>0.
 	\end{aligned}
 \end{equation} 
 Algorithm~\ref{Algo:algo1} is guaranteed to converge to a local minimum with probability one~\cite{spall2005introduction}. 
 We consider the following stopping criteria:  $|| \hat{\nabla}_{\theta_{\widetilde{D}}} \widetilde{H}(\theta_{\widetilde{D},t})||_2 \leqslant  \varepsilon$, or  algorithm stops when it reaches a user--defined maximum number of iterations. 
 Finally, $\widetilde{H}(.)$ in Eq.~(\ref{eq:modified_cost}) is estimated using Function~\ref{Algo:algo2}.

 \begin{algorithm}[h!]
 	\floatname{algorithm}{Function}	 
 	\caption{$\widetilde{H}(.)$}
 	\begin{algorithmic}[1]
 		\renewcommand{\algorithmicrequire}{\textbf{Require:}}
 		\renewcommand{\algorithmicensure}{\textbf{Output:}}		
 		\REQUIRE parameter $\theta_{\widetilde{D}}$ and $\pi_0$
 		\ENSURE $\widetilde{H}(\theta_{\widetilde{D}})$
 		\\ \textit{Initialization} : $k=0$ and $\widetilde{H} =0$
 		\WHILE{$\theta_{\widetilde{D}_k}^T \pi_{k} \geqslant 0$}
 		\STATE $k = k+1$
 		\STATE Obtain a new feature $F_k$
 		\STATE Update $\pi_{k}$  using Eq.~(\ref{eq:update_rule})		
 		\STATE $\widetilde{H} = \widetilde{H} + c_{k}$
 		\ENDWHILE
 		\STATE $\widetilde{H} = \widetilde{H} + M_{\widetilde{D}_k}^T\pi_{k}$
 		\RETURN $\widetilde{H}$
 	\end{algorithmic}
 	\label{Algo:algo2}
 \end{algorithm}

%

\section{Experimental Results}\label{sec:Exp}
\begin{table}[t]
	\centering
	\caption{Datasets used in our experiments.}
		\vspace{-0.1in}
	\label{tbl:dataset}
	\begin{footnotesize}
\begin{tabular}{ |p{1.0cm}|p{1.4cm}|p{1.3cm}|p{1.1cm}|  }
	\hline
	Dataset   & \# Instances  & \# Features &   \# classes                \\   \hline
	Madelon  &   $2,000$      &$500 $   &   $2$        \\   \hline
	Lung  &   $181$      &$12,533$     &   $2$        \\  \hline
	MLL &   $72$      &$5,848$     &   $3$        \\  \hline
	Dexter    &   $300$      &$20,000$     &   $2$        \\  \hline
	Car    &   $174$      &$9,182$     &   $11$        \\  \hline
	Lung2    &   $203$      &$3,312$     &   $5$        \\  \hline
	News20   &   $19,996$      &$1,355,191$     &   $2$        \\  \hline
\end{tabular}
	\end{footnotesize}
\end{table}

\begin{figure*}[t]
	\centering
	\includegraphics[width=\textwidth]{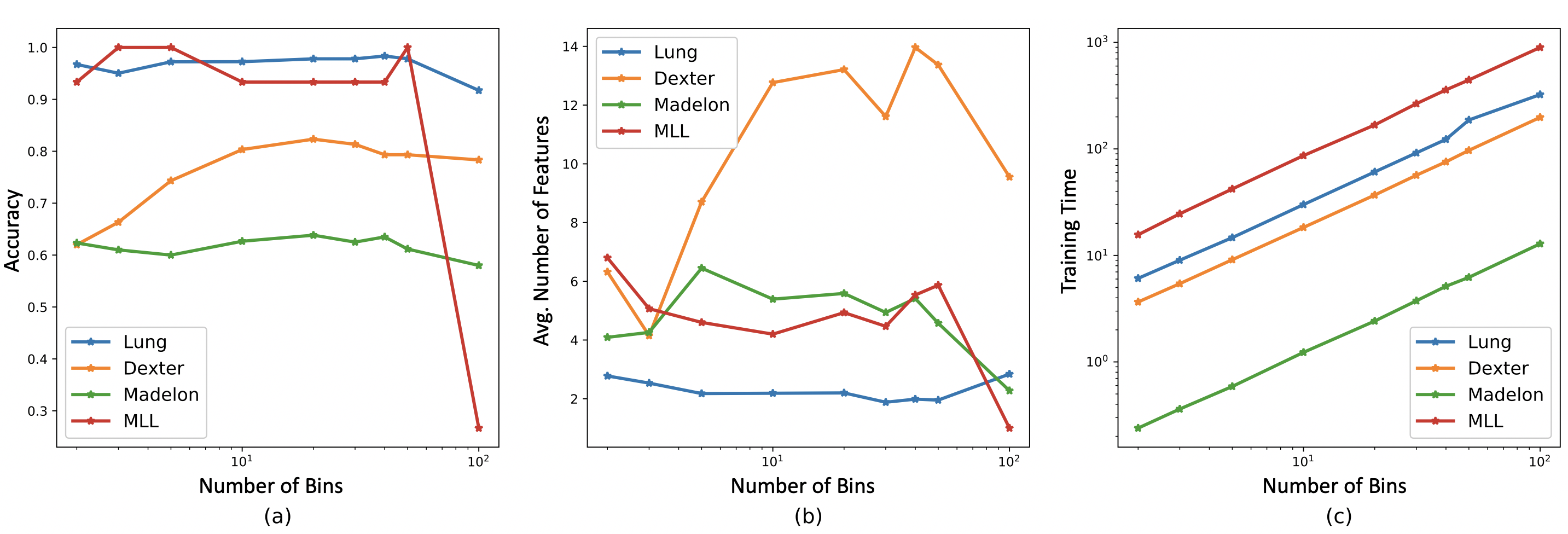} 
	\vspace{-0.3in}
	\caption{Variation of (a) accuracy, (b) average number of features, and (c) training time (sec) as a function of the number $V \in \{2, 3, 5, 10, 20, 30, 40, 50, 100\}$ of bins using Lung Cancer, Dexter, Madelon and MLL datasets.}
	\label{fig:all_bin}
\end{figure*}


In this section, we conduct an extensive set of experiments to evaluate the performance of ETANA and F--ETANA using seven benchmark datasets: 4 DNA Microarray Datasets (Lung Cancer, Lung2, MLL, Car)  \cite{yang2006stable}, 2 NIPS 2003 feature selection challenge datasets (Dexter, Madelon) \cite{NIPSdataset}, and 1 high dimensional dataset (News20)~\cite{Libsvm}. Table~\ref{tbl:dataset} summarises these datasets. For Madelon, Dexter and MLL datasets, we use the originally provided training and validation sets, while for Lung Cancer, Lung2 and Car datasets, we report five--fold cross validated results. All experiments are conducted on a PC with Intel(R) Core(TM) i7-7700 @3.60 GHz CPU with 16 GB memory, running Windows 10 Pro, 64 bit operating system.

\subsection{Practical Considerations} \label{subsec:OS_prac}
Here, we discuss some practical considerations. We use a smoothed maximum likelihood estimator to estimate $p(F_k|T_i), k = 1, \dotsc, K, i = 1,\dots,N$, after quantizing the feature space. Specifically, $\hat{p}(F_k|T_i) = \frac{S_{k,i} + 1}{S_i + V}$, where $S_{k,i}$ denotes the number of samples that satisfy $F_{k} = f_k$ and belong to class $T_i$, $S_i$ denotes the total number of samples belonging to class $T_i$, and $V$ is the number of bins considered.  The effect of the number $V$ of bins on the performance of our algorithm is studied in Section \ref{sec:er_bin}. We estimate the \emph{a priori} probabilities as $P(T_i)= \frac{S_i}{\sum_{i=1}^{N}S_i},  i = 1, \dotsc, N$.

Feature ordering is crucial for early stopping. Different features can hinder or facilitate the quick identification of the class of which an instance may belong to. Consider an example of classifying fruits as either `Apple' or `Orange' using two features $F_1$ and $F_2$, where $F_1$ is the color of the fruit, and $F_2$ is the weight of the fruit. Intuitively, the color of the fruit can potentially simplify the classification process as compared to the weight of the fruit. As a result, if feature $F_2$ was to be examined first, it would be very probable for feature $F_1$ to be examined as well to improve the chances of accurate classification. Instead, if $F_1$ was to be evaluated first, a decision could be made using one feature only. To avoid the computational complexity of evaluating all $K!$ possible feature orderings, we sort features in increasing order of the sum of type I and II errors (considering the true class as the positive class and all the rest classes as a single negative class), scaled by the cost coefficient of the $n$th feature to promote low cost features that at the same time are expected to result in few errors. Finally, for F--ETANA, we set $\varepsilon=10^{-5}$, and $t_{\text{max}}=10^{5}$ as the stopping criteria in Algorithm~\ref{Algo:algo1}. 

\subsection{Effect of Feature Space Quantization}\label{sec:er_bin}

In Section \ref{subsec:OS_prac}, we estimated the conditional probabilities of features given the class using a data binning technique (i.e., $\hat{p}(F_k|T_i) = \frac{N_{n,i} + 1}{N_i + V}$). In this subsection, we analyze the effect of the number $V$ of bins on ETANA using four datasets (Lung, Dexter, Madelon, MLL). 

In Fig.~\ref{fig:all_bin}, we plot the variation of the accuracy, the average number of features used for classification, and the training time as a function of $V$. ETANA's accuracy and the average number of features used for classification is relatively robust to the number of bins (see Fig.~\ref{fig:all_bin}(a) and Fig.~\ref{fig:all_bin}(b)). However, increasing the number of bins from 50 to 100 results in a drop in accuracy for the MLL dataset. Most probably, this is due to overfitting as a result of increasing the resolution of the feature space to a very high value. 
The linear relationship between training time and the number of bins (see Fig.~\ref{fig:all_bin}(c)) is due to Eq.~(\ref{eq:dp_programming}). Without any loss of generality, in the rest of the experiments we set $V$ to the number of class variables (i.e., $N$) in the evaluating dataset.  

\begin{table*}[t]
	\centering
	\caption{Comparison of accuracy. The highest accuracy is bolded and gray--shaded. The second highest value is also gray shaded. Cells are marked with `- -' if the corresponding method was unable to generate results within a cutoff time of 12 days.}
	\vspace{-0.1in}
	\label{tbl:acc}
	\begin{footnotesize}
	\begin{tabular}{|p{1.0cm}|p{0.8cm}|p{1.3cm}|p{1.7cm}|p{01.3cm}|p{0.8cm}|p{1.4cm}|p{0.8cm}|p{2.1cm}|}
		\hline
		Dataset &ETANA &F--ETANA   & OFS--Density & OFS--A3M & SAOLA & Fast--OSFS & OSFS & Alpha--Investing \\
		\hline
		Madelon  &\cellcolor{gray!20}  $\textbf{0.6233}$ &$0.555$     &$0.5117$    &$0.5117$      &$0.5817$ &$0.5417$ &$0.5817$        &\cellcolor{gray!20} $0.6050$   \\   \hline
		Lung C.  &$0.9673$  &\cellcolor{gray!20}  $0.9835$     &$0.9779$    &$0.9557$      &\cellcolor{gray!20}  $\textbf{0.9890}$ &\cellcolor{gray!20}  $\textbf{0.9890}$ &$0.9724$        &$0.9613$    \\   \hline
		MLL  &\cellcolor{gray!20}  $\textbf{1.00}$  &\cellcolor{gray!20}  $\textbf{1.00}$   &\cellcolor{gray!20} $0.9333$  &$0.8667$      &$0.8667$ &$0.8000$ &$0.8000$        &\cellcolor{gray!20} $0.9333$    \\   \hline
		Dexter &$0.62$  &$0.62$     &\cellcolor{gray!20}  $\textbf{0.8167}$    &$0.7800$      &$0.7800$  &$0.7800$  &\cellcolor{gray!20} $0.7967$        &$0.5000$    \\   \hline
		Car  &\cellcolor{gray!20} $0.8217$ &\cellcolor{gray!20}  $\textbf{0.8273}$    &$0.5854$    &$0.7874$      &$0.7982$ &$0.5869$ &$0.500$        &$0.6429$    \\   \hline
		Lung2  &$0.8720$  &$0.8818$     &\cellcolor{gray!20}  $\textbf{0.8968}$    &\cellcolor{gray!20} $0.8865$    &$0.8817$ &$0.8420$ &$0.8471$        &$0.8820$    \\   \hline
		News20 &\cellcolor{gray!20} $0.7352$  &$0.6346$     &$--$    &\cellcolor{gray!20}  $\textbf{0.7846}$      &$--$ &$--$ &$--$        &$--$    \\   \hline
	\end{tabular}
\end{footnotesize}
\end{table*}

\begin{table*}[t]
	\centering
	\caption{Comparison of average number of features used. Values corresponding to highest and second highest accuracy are bolded and gray--shaded, and gray--shaded accordingly. Cells are marked with `- -' if the corresponding method was unable to generate results within a cutoff time of 12 days.}
	\vspace{-0.1in}
	\label{tbl:feat}
	\begin{footnotesize}
	\begin{tabular}{|p{1.0cm}|p{0.8cm}|p{1.3cm}|p{1.7cm}|p{01.3cm}|p{0.8cm}|p{1.4cm}|p{0.8cm}|p{2.1cm}|}
		\hline
		Dataset &ETANA &F--ETANA   & OFS--Density & OFS--A3M & SAOLA & Fast--OSFS & OSFS & Alpha--Investing \\ \hline
		Madelon  &\cellcolor{gray!20}  $\textbf{4.09}$  &$211.21$   &$2$    &$2$      &$3$ &$3$ &$3$        &\cellcolor{gray!20} $4$    \\   \hline
		Lung C.  &$2.78$ &\cellcolor{gray!20} $8.48$     &$25.40$    &$8.20$      &\cellcolor{gray!20}  $\textbf{52}$ &\cellcolor{gray!20}  $\textbf{6.8}$ &$4.0$        &$4.6$    \\   \hline
		MLL    &\cellcolor{gray!20}  $\textbf{5.07}$  &\cellcolor{gray!20}  $\textbf{10.8}$     &\cellcolor{gray!20} $12$    &$9$      &$28$ &$5$ &$3$        &$7$    \\   \hline
		Dexter   &$6.32$ &$7.04$      &\cellcolor{gray!20}  $\textbf{10}$    &$98$      &$21$ &$9$ &\cellcolor{gray!20} $6$        &$1$    \\   \hline
		Car   &\cellcolor{gray!20} $13.85$   &\cellcolor{gray!20}  $\textbf{289.38}$   &$34.8$    &$36.6$      &$41.40$ &$8.4$ &$4.4$        &$24.4$    \\   \hline
		Lung2    &$13.77$ &$71.66$      &\cellcolor{gray!20}  $\textbf{14.2}$    &\cellcolor{gray!20} $18.4$     &$28.2$ &$9.4$ &$5.8$        &$34.4$    \\   \hline
		News20  &\cellcolor{gray!20} $81.70$ &$4000.6$     &$--$    &\cellcolor{gray!20}  $\textbf{241.8}$      &$--$ &$--$ &$--$        &$--$    \\   \hline
	\end{tabular}
\end{footnotesize}
\end{table*}

\begin{table*}[t]
	\centering
	\caption{Comparison of time (in seconds) required for feature selection (F), classification (C), joint feature selection and classification (F+C), and model training (T). Values corresponding to highest and second highest accuracy are bolded and gray--shaded, and gray--shaded accordingly. Cells are marked with `- -' if the corresponding method was unable to generate results within a cutoff time of 12 days.}
	\vspace{-0.1in}
	\label{tbl:time}
		\begin{footnotesize}
				\begin{tabular}{|p{1.0cm}|p{0.1cm}|p{0.8cm}|p{1.3cm}|p{0.1cm}|p{1.7cm}|p{01.3cm}|p{0.8cm}|p{1.4cm}|p{0.8cm}|p{2.1cm}|}
		\hline
		Dataset & \rotatebox[origin=c]{90}{Time }  &ETANA &F--ETANA  & \rotatebox[origin=c]{90}{Time }  & OFS--Density & OFS--A3M & SAOLA & Fast--OSFS & OSFS &Alpha--Investing \\
		\hline	
		\multirow{3}{*}{Madelon}
		&\multirow{2}{*}{\parbox{0.2cm}{\centering{F+ C}}}  &\cellcolor{gray!20}      &$\multirow{2}{*}{3.678}$   &F    &$188.85$    &$225.48$      &$0.136$ &$0.199$ &$0.203$        &\cellcolor{gray!20} $0.043$    \\    \cline{5-11} 
		& & \cellcolor{gray!20}  $\multirow{-2}{*}{\textbf{0.062}}$  &   &C    &$0.024$    &$0.025$      &$0.025$ &$0.025$ &$0.026$        &\cellcolor{gray!20} $0.024$    \\   
		\cline{2-11}
		&  T   &\cellcolor{gray!20}  $\textbf{0.141}$  &$0.34$   &T    &$0.423$    &$0.424$      &$0.438$ &$0.429$ &$0.437$        &\cellcolor{gray!20} $0.438$   \\   \hline

		\multirow{3}{*}{\parbox{1.2cm}{\centering Lung Cancer}}
		&\multirow{2}{*}{\parbox{0.2cm}{\centering{F+ C}}} &$\multirow{2}{*}{0.003}$   &\cellcolor{gray!20}   &F     &$13.455$    &$39.031$      &\cellcolor{gray!20}  $\textbf{4.692}$ &\cellcolor{gray!20}  $\textbf{2.644}$ &$37.868$        &$1.207$    \\   \cline{5-11}
		&  &   &  \cellcolor{gray!20} \multirow{-2}{*}{$0.014$}   &C     &$0.017$    &$0.019$      &\cellcolor{gray!20}  $\textbf{0.017}$ &\cellcolor{gray!20}  $\textbf{0.012}$ &$0.012$        &$0.012$    \\ 
		\cline{2-11}
		&  T &$3.680$   &\cellcolor{gray!20} $2.392$   &T     &$0.117$    &$0.114$      &\cellcolor{gray!20}  $\textbf{0.118}$ &\cellcolor{gray!20}  $\textbf{0.120}$ &$0.119$        &$0.121$    \\   \hline

		\multirow{3}{*}{MLL}
		&\multirow{2}{*}{\parbox{0.2cm}{\centering{F+ C}}}  &\cellcolor{gray!20}  & \cellcolor{gray!20}   &F   &\cellcolor{gray!20} $1.524$   &$5.122$      &$3.079$ &$1.495$ &$9.354$        &$0.463$    \\   \cline{5-11}
		&  &  \cellcolor{gray!20}  $\multirow{-2}{*}{\textbf{0.001}}$   &  \cellcolor{gray!20} $\multirow{-2}{*}{\textbf{0.001}}$     &C    &\cellcolor{gray!20} $0.045$    &$0.023$      &$0.045$ &$0.025$ &$0.025$        &$0.023$    \\  
		\cline{2-11}
		& T &\cellcolor{gray!20}  $\textbf{15.376}$   &\cellcolor{gray!20}  $\textbf{2.28}$     &T   &\cellcolor{gray!20} $0.397$    &$0.401$      &$0.408$ &$0.412$ &$0.409$        &$0.429$    \\   \hline

		\multirow{3}{*}{Dexter}
		&\multirow{2}{*}{\parbox{0.2cm}{\centering{F+ C}}}   &$\multirow{2}{*}{0.047}$  &$\multirow{2}{*}{0.046}$   &F    &\cellcolor{gray!20}  $\textbf{98.482}$    &$43554$      &$1.006$ &$1.513$ &\cellcolor{gray!20} $3.195$        &$15.663$    \\   \cline{5-11}
		& &  &   & C     &\cellcolor{gray!20}  $\textbf{0.072}$    &$0.080$      &$0.085$ &$0.072$ &\cellcolor{gray!20} $0.066$  &$0.054$    \\   \cline{2-11}
		&T &$2.484$  &$2.24$   &T    &\cellcolor{gray!20}  $\textbf{0.420}$    &$0.423$      &$0.420$ &$0.423$ &\cellcolor{gray!20} $0.427$        &$0.424$    \\   \hline

		\multirow{3}{*}{Car}
		&\multirow{2}{*}{\parbox{0.2cm}{\centering{F+ C}}}  &\cellcolor{gray!20}   &\cellcolor{gray!20}   &F    &$7.882$    &$37.205$      &$2.249$ &$1.092$ &$14.144$        &$0.956$    \\   \cline{5-11}
		& &   \cellcolor{gray!20}   \multirow{-2}{*}{$0.037$} & \cellcolor{gray!20}  $\multirow{-2}{*}{\textbf{0.332}}$ &C     &$0.012$    &$0.017$      &$0.017$ &$0.001$ &$0.001$        &$0.017$    \\    \cline{2-11}
		&T  &\cellcolor{gray!20} $1167.5$ &\cellcolor{gray!20}  $\textbf{36.743}$  &T     &$0.113$    &$0.112$      &$0.117$ &$0.004$ &$0.005$        &$0.117$    \\   \hline

		\multirow{3}{*}{Lung2}
		&\multirow{2}{*}{\parbox{0.2cm}{\centering{F+ C}}}  &$\multirow{2}{*}{0.034}$    &$\multirow{2}{*}{0.071}$ &F     &\cellcolor{gray!20}  $\textbf{4.214}$    &\cellcolor{gray!20} $16.346$     &$1.255$ &$1.681$ &$28.888$        &$0.338$    \\   \cline{5-11}
		& &     & &C     &\cellcolor{gray!20}  $\textbf{0.017}$    &\cellcolor{gray!20} $0.017$      &$0.017$ &$0.017$ &$0.012$        &$0.017$    \\   \cline{2-11}
		&T    &$715.03$    &$7.713$ &T     &\cellcolor{gray!20}  $\textbf{0.113}$    &\cellcolor{gray!20} $0.109$      &$0.115$ &$0.120$ &$0.119$        &$0.116$    \\   \hline
		
		\multirow{3}{*}{News20}
		&\multirow{2}{*}{\parbox{0.2cm}{\centering{F+ C}}}  &\cellcolor{gray!20}     &$\multirow{2}{*}{346.47}$ &F     &$--$    &\cellcolor{gray!20}  $\textbf{2183}$      &$--$ &$--$ &$--$        &$--$    \\   \cline{5-11}
		& & \cellcolor{gray!20}     \multirow{-2}{*}{$117.61$}  & &C     &$--$    &\cellcolor{gray!20}  $\textbf{60.60}$      &$--$ &$--$ &$--$        &$--$    \\  \hhline{~-|-|-|-|-|-|-|-|-|-|} 
		&T   &\cellcolor{gray!20}    $3076$    &$429.7$ &T     &$--$    &\cellcolor{gray!20}  $\textbf{0.586}$      &$--$ &$--$ &$--$        &$--$    \\   \hline
		
	\end{tabular}
	\end{footnotesize}
\end{table*}

\subsection{Accuracy as a Function of the Average Number of Features Used}

To study the behavior of ETANA for varying values of feature evaluation cost $c$, when all features incur same cost (i.e., $c_k =c$), we measured accuracy for constant misclassification costs (i.e., $M_i,j =1 \forall i \neq j, M_{i,i} = 0, i,j\in \{1,\dots,N\}$) and $c = \{0.1, 0.08, 0.06, 0.04, 0.02, 0.01, 0.001, 0\}$. Different $c$ values result in different number of features and levels of accuracy. 
Intuitively, using a small potion of the total feature set leads to low accuracy, whereas when the average number of features used increases, the performance improves dramatically. From here onwards, unless specified, we report results for $c=0.01$.

\subsection{Comparison with Online Feature Selection Methods}
	
In this subsection, we compare ETANA and F--ETANA with the following state--of--the--art feature selection methods: OFS--Density~\cite{zhou2019ofs}, OFS--A3M~\cite{zhou2019online}, SOAOLA~\cite{yu2014towards}, OSFS~\cite{wu2012online}, Fast--OSFS\cite{wu2012online}, and Alpha--Investing~\cite{zhou2005streaming}. We use KNN classifier with three neighbours to evaluate a selected feature subset, which has been shown to outperform SVM, CART, and J48 classifiers on the datasets used in~\cite{zhou2019ofs,yu2014towards}. For SAOLA, OSFS, and Fast--OSFS,  the parameter $\alpha$ is set to $0.01$~\cite{yu2014towards,wu2012online}. For Alpha--Investing, parameters are set to the values used in~\cite{zhou2005streaming}. 

We summarize our observations from Tables 2, 3, 4 by dataset as follows. 

\noindent\textbf{Madelon:} ETANA achieves the highest accuracy using only $\sim$4 features on average. In fact, ETANA achieves an improvement of $3\%$ in accuracy over Alpha--Investing, which has the highest accuracy among all the baselines using the same number of features. At the same time, ETANA is $7.5\%$, and $67.8\%$ faster in joint feature selection and classification, and model training respectively, compared to Alpha--Investing. \\
\textbf{Lung Cancer:} SAOLA and Fast--OSFS achieve the highest accuracy using $52$, and $6.8$ features, respectively, but require  $18.7$, and $2.4$ times more features respectively, compared to ETANA for a difference of  $2\%$ in accuracy. Further, ETANA is much faster in joint feature selection and classification compared to SAOLA and Fast--OSFS. \\
\textbf{MLL:} ETANA and F--ETANA achieve  $100\%$ accuracy using $5.07$, and $10.8$ features on average, respectively. This corresponds to an improvement of $7\%$ in accuracy with  $57.8\%$, and $27.6\%$  less of features used compared to  OFS--Density and Alpha--Investing, respectively, which achieve the highest accuracy among all the baselines. At the same time, both ETANA and F--ETANA are much faster in joint feature selection and classification compared to OFS--Density and Alpha--Investing. \\
\textbf{Dexter:} OFS--Density achieves the highest accuracy, but requires $58.2\%$ more features. This results in a significant slowdown for joint feature selection and classification compared to ETANA. \\
\textbf{Car:} F--ETANA achieves the highest accuracy, however, ETANA is a close second while using only 13.85 features on average. This corresponds to an improvement of $3\%$ and $66.5\%$ in accuracy and average number of features used respectively, while at the same time, leads to a faster runtime compared to SAOLA, the best performing baseline. \\
\textbf{Lung2:}  OFS--Density achieves the highest accuracy with $3.1\%$ more features as compared to ETANA, and much slower runtime.


\subsection{Performance Assessment on a High Dimensional Dataset}

In this subsection, we discuss the performance of our algorithms, ETANA and F--ETANA, and the state--of--the--art online feature selection methods on the News20 dataset. Experiments on this dataset are conducted using the high performance computing cluster provided by the Information Technology Services at the University of Albany, SUNY. We used one node with 20  Intel(R) Xeon(R) E5-2680 v4 @2.40GHz CPUs with 256 GB memory. Except for SAOLA, the rest of the online feature selection methods were unable to generate results within a cutoff time of 12 days. 
Although SAOLA achieves the highest accuracy, it requires $\sim$200\% more features and is $\sim$20 times slower in joint feature selection and classification  compared to ETANA for $c=0.001$ (see the last row in  Tables 2, 3, 4). 

\subsection{F--ETANA versus ETANA}

\begin{figure}[t]
	\centering
	\includegraphics[width=\columnwidth]{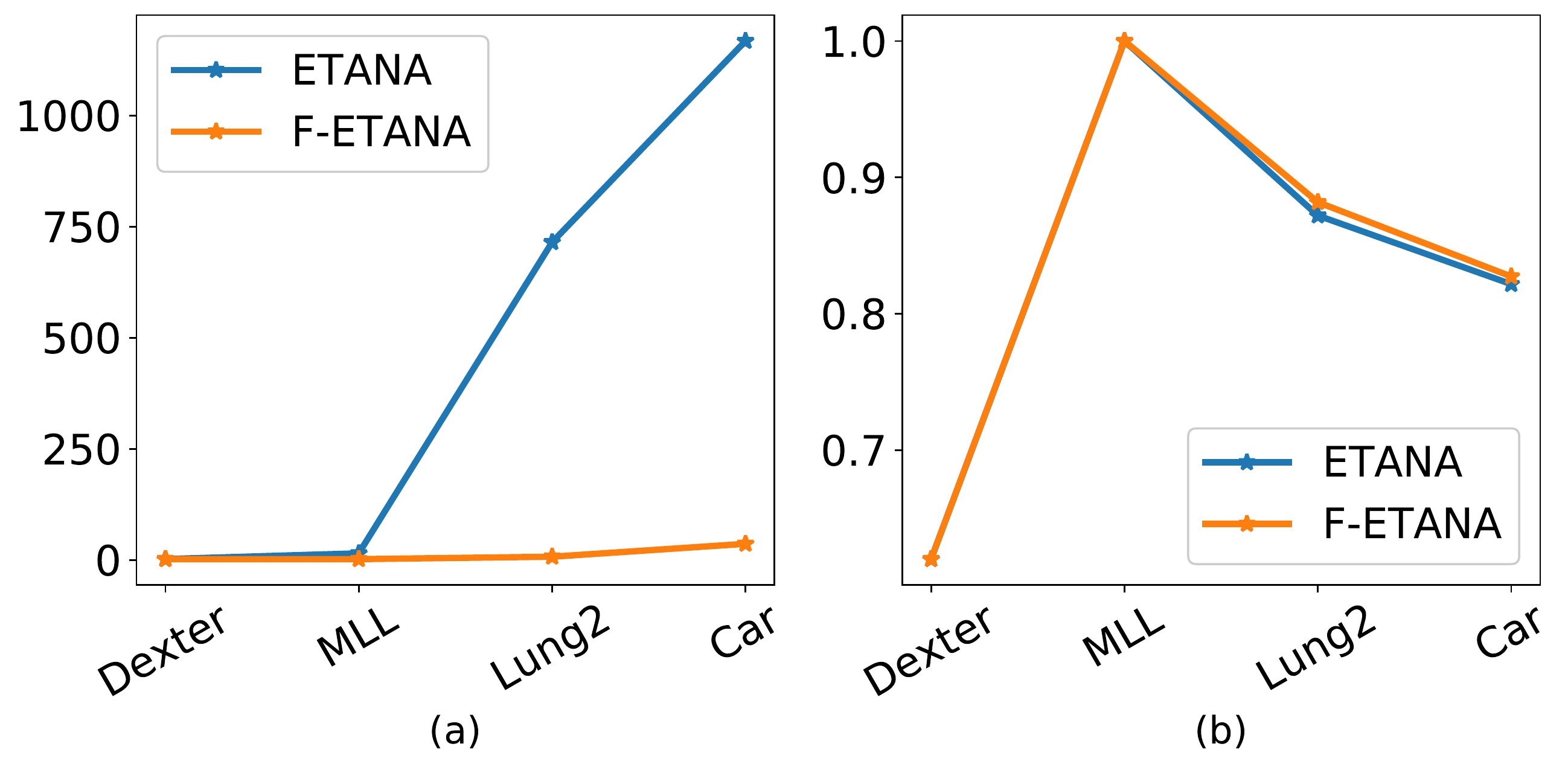} 
	\vspace{-0.2in}
	\caption{Comparison of (a) model training time (seconds), and (b)  accuracy achieved by F--ETANA and ETANA using  Dexter (2 classes), MLL (3 classes), Lung2 (5 classes), and CAR datasets (11 classes).  }
	\label{fig:fast_opot}
	\vspace{-0.2in}
\end{figure}

Thus far we have shown that ETANA  outperforms all baselines in terms of accuracy, number of features, and time required for joint feature selection and classification. The limitation of ETANA is in its training time  (see Table~\ref{tbl:time}),  due to the construction of a $ (K+1) \times d$ matrix which grows exponentially with the number of classes (see Section~\ref{sec:PF}).  F--ETANA drastically reduces the time required for model training as compared to ETANA  without sacrificing accuracy (see  Fig.~\ref{fig:fast_opot}). At the same time, F--ETANA requires more features per data instance compared to ETANA (see Table~\ref{tbl:feat}).

\section{Conclusion}

This paper investigated a new research problem, on--the--fly joint feature selection and classification, which aims to minimize the number of feature evaluations per data instance for fast and accurate classification. Specifically, an optimization problem was defined in terms of the cost of evaluating features and the Bayes risk associated with the classification decision. The optimum solution was derived using dynamic programming and it was shown that the corresponding functions are concave, continuous and piecewise linear.  Two algorithms, ETANA and F--ETANA were proposed based on the optimum solution and its properties. The proposed algorithms outperformed state--of--the--art feature selection methods in terms of the average number of features used, classification accuracy, and the time required for on--the--fly joint feature selection and classification.  Furthermore, F--ETANA resulted in a drastic reduction in model training time compared to ETANA.  As a part of our future work, we plan to exploit feature dependencies, which may improve performance even more. 
  
\appendices
\section{}
\subsection{Proof of Lemma~\ref{lem:posterior_probability}}
We start from the definition of the \textit{a posteriori probability} vector, i.e.:
\begin{align*}
\pi_{k} &\triangleq [\pi_{k}^1,\pi_{k}^2,\dots,\pi_{k}^N]^T,
\end{align*}
and consider any element in this vector, i.e., $\pi_{k}^i$. Specifically, we use Bayes' rule and the law of total probability to get the following result:
\begin{align}
\pi_{k}^i &= P(T_i | F_{1}, \dotsc, F_{k}) \nonumber \\
\label{eq.posteriori2}
&= \frac{P(F_{1}, \dotsc, F_{k}| T_i) P(T_i)}{P(F_{1}, \dotsc, F_{k})}  \\
&= \frac{P(F_{1}, \dotsc, F_{k}| T_i) P(T_i)}{\sum_{j = 1}^N P(F_{1}, \dotsc, F_{k},T_j)} \nonumber \\
\label{eq.posteriori3}
&= \frac{P(F_{1}, \dotsc, F_{k}| T_i) P(T_i)}{\sum_{j = 1}^N P(F_{1}, \dotsc, F_{k} | T_j)P(T_j)}. 
\end{align}
Note that we can further simplify Eq.~(\ref{eq.posteriori3}) by exploiting the conditional independence of the features in set $F$ given the class variable $T$ as follows:
\begin{align}
\pi_{k}^i &=  \frac{ P(T_i) \prod_{n=1}^{k}P(F_{n}| T_i)}{\sum_{j = 1}^N P(T_j) \prod_{n=1}^{k}P(F_{n}| T_j)} \nonumber \\
\label{eq.posteriori1}
&=  \frac{ p_i \prod_{n=1}^{k}P(F_{n}| T_i)}{\sum_{j = 1}^N p_j \prod_{n=1}^{k}P(F_{n}| T_j)}.
\end{align}
Similarly, $\pi_{k-1}^i$ will take the following form:
\begin{align}
\label{eq.posteriori4}
\pi_{k-1}^i =  \frac{ p_i \prod_{n=1}^{k-1}P(F_{n}| T_i)}{\sum_{j = 1}^N p_j \prod_{n=1}^{k-1}P(F_{n}| T_j)}.
\end{align}
We can now rewrite $\pi_{k}^i$ in Eq.~(\ref{eq.posteriori1}) in terms of $\pi_{k-1}^i$ in Eq.~ (\ref{eq.posteriori4}) as follows:
\begin{align}
\pi_{k}^i =& \frac{ p_i \prod_{n=1}^{k}P(F_{n}| T_i)}{\sum_{j = 1}^N p_j \prod_{n=1}^{k}P(F_{n}| T_j)} \nonumber \\
&= \frac{ P(F_{k}| T_i) \big( p_i \prod_{n=1}^{k-1}P(F_{n}| T_i) \big)}{\sum_{j = 1}^N p_j \prod_{n=1}^{k}P(F_{n}| T_j)} \nonumber \\
&= \frac{ P(F_{k}| T_i) \big( \pi_{k-1}^i \sum_{j = 1}^N p_j \prod_{n=1}^{k-1}P(F_{n}| T_j)  \big)}{\sum_{j = 1}^N p_j \prod_{n=1}^{k}P(F_{n}| T_j)} \nonumber \\
&= \frac{ P(F_{k}| T_i) \pi_{k-1}^i   }{ \frac{\sum_{j = 1}^N p_j \prod_{n=1}^{k}P(F_{n}| T_j)}{\sum_{j = 1}^N p_j \prod_{n=1}^{k-1}P(F_{n}| T_j)}} \nonumber \\
&= \frac{ P(F_{k}| T_i) \pi_{k-1}^i   }{ \sum_{j = 1}^N  P(F_{k}| T_j) \pi_{k-1}^j}. 
\end{align}
Finally, using the above result, the \textit{a posteriori probability} vector takes the following form:
\begin{align}
\pi_{k} &= [\pi_{k}^1,\pi_{k}^2,\dots,\pi_{k}^N]^T \nonumber \\
&= \frac{ \diag([P(F_{k}| T_1),\dots,P(F_{k}| T_N)]) [\pi_{k-1}^1,\dots,\pi_{k-1}^N]^T}{ [P(F_{k}| T_1),\dots,P(F_{k}| T_N)] [\pi_{k-1}^1,\dots,\pi_{k-1}^N]^T  } \nonumber \\
&= \frac{\diag([P(F_{k}| T_1),\dots,P(F_{k}| T_N)]) \pi_{k-1} }{[P(F_{k}| T_1),\dots,P(F_{k}| T_N)] \pi_{k-1} } \nonumber \\
&= \frac{\diag(\Delta_k(F_k)) \pi_{k-1}}{\Delta_k^T(F_k) \pi_{k-1}},
\end{align}
where $\Delta_k(F_k) \triangleq [P(F_{k}| T_1),P(F_{k}| T_2),\dots,P(F_{k}| T_N)]^T$,  $\diag(A)$ denotes a diagonal matrix with diagonal elements being the elements in vector $A$, and $\pi_{0} \triangleq [p_1,p_2,\dots,p_N]^T$.
\subsection{Proof of Lemma~\ref{lem:indicator}}
Using the law of total probability, we can write the probability $P(D_{R} = j,T_{i})$ as follows:
\begin{align}
P(D_{R} = j,T_{i}) &= \sum_{k=1}^{K} P(R = k,D_{k} = j,T_{i}) \nonumber \\
\label{eq.indicator5}
&= \sum_{k=1}^{K} P(T_{i}) P(R = k,D_{k} = j|T_{i}). 
\end{align}
Using the fact that the event ${\lbrace R = k, D_{k} = j \rbrace}$ depends only on the set $\lbrace F_1, F_2, \dotsc, \allowbreak F_k \rbrace$, and by the definition $P(R = k, D_k = j ) = E \lbrace \mathbbm{1}_{\lbrace R = k \rbrace} \mathbbm{1}_{\lbrace D_{k} = j \rbrace} \rbrace$, Eq.~(\ref{eq.indicator5}) can be written as follows:
\begin{align}
\label{eq.indicator1}
P(D_{R} = j,T_{i}) = &\sum_{k=1}^{K}  \sum_{F_1,F_2,\dots,F_k}  \mathbbm{1}_{\lbrace R = k \rbrace} \mathbbm{1}_{\lbrace D_{k} = j \rbrace}  \nonumber \\ 
& \times P(T_{i}) P(F_{1}, \dotsc, F_{k}| T_i). 
\end{align}
Then, using the result in Eq.~(\ref{eq.posteriori2}), we can incorporate  $\pi_{k}^i$ in Eq.~(\ref{eq.indicator1}) as follows:
\begin{align}\label{eq.indicator2}
P(D_{R} = j,T_{i})  = & \sum_{k=1}^{K}  \sum_{F_1,F_2,\dots,F_k}  \mathbbm{1}_{\lbrace R = k \rbrace} \mathbbm{1}_{\lbrace D_{k} = j \rbrace}  \nonumber \\ 
&\times P(F_{1}, \dotsc, F_{k}) \pi_{k}^i. 
\end{align}
Further, from the definition of the expectation operator (i.e., if a random variable $Y$ has set $D$ of possible values and probability mass function $P(Y)$, then the expected value $\mathbb{E}\lbrace H(Y) \rbrace$ of any function $H(Y)$ equals $\sum_{Y\in D} P(Y)H(Y) $), Eq.~(\ref{eq.indicator2}) can be rewritten as follows:
\begin{align}\label{eq.indicator3}
P(D_{R} = j,T_{i})  &= \sum_{k=1}^{K}   \mathbb{E}  \big \lbrace   \mathbbm{1}_{\lbrace R = k \rbrace} \mathbbm{1}_{\lbrace D_{k} = j \rbrace}  \pi_{k}^i \big \rbrace. 
\end{align}
By the \textit{linearity of expectation} in Eq.~(\ref{eq.indicator3}), we get:
\begin{align}\label{eq.indicator4}
P(D_{R} = j,T_{i}) &= \mathbb{E} \bigg \lbrace \sum_{k=1}^{K}        \mathbbm{1}_{\lbrace R = k \rbrace} \mathbbm{1}_{\lbrace D_{k} = j \rbrace}  \pi_{k}^i \bigg \rbrace. 
\end{align}
Finally, using the fact that $x_{R} = \sum_{k = 0}^{K} x_{k} \mathbbm{1}_{\lbrace R = k \rbrace}$,  Eq.~(\ref{eq.indicator4}) will end up in the desired form as shown below:
\begin{align}
P(D_{R} = j,T_{i}) &= \mathbb{E} \big \lbrace  \pi_R^i  \mathbbm{1}_{\lbrace D_{R} = j \rbrace}  \big \rbrace. 
\end{align}

\subsection{Proof of Theorem~\ref{thm:optimal_selection_strategy}}

At any stopping time $R$, the optimum decision $D_R$ takes only one out of $N$ possibilities such that the misclassification cost is minimum. In the process of finding this optimum $D_R$, it is important to note that $\sum_{j=1}^{N} \mathbbm{1}_{\lbrace D_{R} = j \rbrace} = 1$, where $\mathbbm{1}_{A}$ is the indicator function for event $A$ (\emph{i.e.,} $\mathbbm{1}_{A} = 1$ when $A$ occurs, and $\mathbbm{1}_{A} = 0$ otherwise), which implies that only one of the terms in the sum becomes $1$, while the remaining terms are equal to zero. We note that:
\begin{align}
M_j^T \pi_R  \geqslant g(\pi_R), \ \forall j \in \{1,2,\dots,N\},
\end{align}
where $g(\pi_{R}) \triangleq \min_{1 \leqslant j \leqslant N} \big [   M_j^T \pi_R \big ]$. Then, using the fact that $\mathbbm{1}_{\lbrace D_{n} = j \rbrace}$ is non--negative, we get the following result:
\begin{align}
\sum_{j=1}^{N} \big(   M_j^T \pi_R \big ) \mathbbm{1}_{\lbrace D_{R} = j \rbrace}  &\geqslant g(\pi_R)  \sum_{j=1}^{N} \mathbbm{1}_{\lbrace D_{R} = j \rbrace} \nonumber \\
&= g(\pi_R).
\end{align}

We underscore that the lower bound $g(\pi_R)$ derived above is independent of the decision $D_R$. Thus, it is obvious that this lower bound can be achieved only by the rule defined in Eq.~(\ref{eq:optimal_classification_strategy}), which is therefore the optimum decision for a given stopping time $R$.

\subsection{Proof of Theorem~\ref{thm:dp_programming}}

At the end of the $K$th stage, assuming that all the features have been examined, the only remaining expected cost is the optimum misclassification cost of selecting among $N$ decision choices at stage $k=K$, which is $\bar{J}_{K}(\pi_{K}) = g(\pi_{K})$ (see Theorem \ref{thm:optimal_selection_strategy}).

Then, consider any intermediate stage $k=0,1,\dots,K-1$. Being at stage $k$, with the available information $\pi_{k}$, the optimum strategy has to choose between, either to terminate the feature evaluation process and incur cost $g(\pi_{k})$, which is the optimum misclassification cost of selecting among $N$ decision choices (see Theorem \ref{thm:optimal_selection_strategy}), or continue and incur cost of $c_{k+1}$ to evaluate feature $F_{k+1}$ and an additional cost $\bar{J}_{k+1}(\pi_{k+1})$ to continue optimally. Thus, the total cost of continuing optimally (referred to as \textit{optimum cost--to--go} \cite{BertsekasDPOC05}) is $c_{k+1}+\bar{J}_{k+1}(\pi_{k+1})$. It is important to note that at stage $k$, we do not know the outcome of examining feature $F_{k+1}$. Thus, we need to consider the expected \textit{optimum cost--to--go}, which is equal to $ c_{k+1}+  \mathbb{E} \big \lbrace\bar{J}_{k+1}(\pi_{k+1})|\pi_{k} \big \rbrace$. Using Lemma \ref{lem:posterior_probability} to express $\pi_{k+1}$ in terms of $\pi_{k}$, and by the definition of the expectation operator (i.e., if a random variable $Y$ has set $D$ of possible values and probability mass function $P(Y)$, then the expected value $\mathbb{E}\lbrace H(Y) \rbrace$ of any function $H(Y)$ equals to $\sum_{Y\in D} P(Y)H(Y) $), we get the \textit{optimum cost--to--go} $\bar{\mathcal{C}}_{k+1}(\pi_{k})$:  
\begin{align}
\bar{\mathcal{C}}_{k+1}(\pi_{k}) &\triangleq c_{k+1}+  \mathbb{E} \big \lbrace\bar{J}_{k+1}(\pi_{k+1})|\pi_{k} \big \rbrace \nonumber \\
\label{eq.dynamic0}
 &= c_{k+1} + \sum_{F_{k+1}} P(F_{k+1}|F_1,F_2,\dots,F_{k}) \nonumber \\
&\times  \bar{J}_{k+1} \bigg (  \frac{\diag \big(\Delta_{k+1}(F_{k+1}) \big) \pi_{k}}{\Delta_{k+1}^T(F_{k+1}) \pi_{k}} \bigg ). 
\end{align}
\noindent Let us simplify the term $P(F_{k+1}|F_1,F_2,\dots,F_{k})$ separately. Specifically, using Bayes' rule and the law of total probability, we observe that:
\begin{align}
P(F_{k+1}|F_1,\dots,F_{k}) &= \frac{P(F_1,\dots,F_{k+1})}{P(F_1,\dots,F_{k})} \nonumber \\
&= \frac{\sum_{j = 1}^N P(F_{1}, \dotsc, F_{k+1}, T_j)}{\sum_{j = 1}^N P(F_{1}, \dotsc, F_{k}, T_j)} \nonumber \\
\label{eq.dynamic3}
&= \frac{\sum_{j = 1}^N P(F_{1}, \dotsc, F_{k+1} | T_j)P(T_j) }{\sum_{j = 1}^N P(F_{1}, \dotsc, F_{k} | T_j)P(T_j)}.	
\end{align}
Note that we can further simplify Eq.~(\ref{eq.dynamic3}) by exploiting the fact that the random variables $F_k$ are independent under each class $T_j$ as follows:
\begin{align}		
P(F_{k+1}|F_1,\dots,F_{k})  &= \frac{\sum_{j = 1}^N P(T_j) \prod_{n=1}^{k+1}P(F_{n}| T_j)}{\sum_{j = 1}^N P(T_j) \prod_{n=1}^{k}P(F_{n}| T_j) }	 \nonumber \\	
\label{eq.dp1}
&= \frac{\sum_{j = 1}^N \bigg(p_j \prod_{n=1}^{k}P(F_{n}| T_j) \bigg)P(F_{k+1}| T_j)}{\sum_{j = 1}^N p_j \prod_{n=1}^{k}P(F_{n}| T_j) }.	
\end{align}

\noindent Using the result in Eq.~(\ref{eq.posteriori1}), we can simplify Eq.~(\ref{eq.dp1}) as follows:
\begin{align}
P(F_{k+1}|F_1,\dots,F_{k})	&= \sum_{j = 1}^N \pi_{k}^j P(F_{k+1}| T_j) \nonumber \\
\label{eq.dynamic2}&= \Delta_{k+1}^T(F_{k+1}) \pi_{k} .
\end{align}
Finally, substituting Eq.~(\ref{eq.dynamic2}) in Eq.~(\ref{eq.dynamic0}), we get the desired result:
\begin{align}
\bar{\mathcal{C}}_{k+1}(\pi_{k}) &=  c_{k+1}+ \sum_{F_{k+1}} \Delta_{k+1}^T(F_{k+1}) \pi_{k} \nonumber \\
& \times  \bar{J}_{k+1} \bigg (  \frac{\diag(\Delta_{k+1}(F_{k+1}))\pi_{k}}{\Delta_{k+1}^T(F_{k+1}) \pi_{k}} \bigg ). 
\end{align}

\section{}

\subsection{Proof of Lemma~\ref{lem:g_function}}
Let us consider the definition of $g(\varpi)$:
\begin{equation*}
g(\varpi)\triangleq \min_{1 \leqslant j \leqslant N} \big [   M_j^T \varpi \big ], \varpi \in [0,1]^{N}.
\end{equation*}

The term $ M_j^T \varpi$ is linear with respect to $\varpi$, and since the minimum of linear functions is a concave, piecewise linear function, we conclude that $g(\varpi)$ is
a concave, piecewise linear function as well. Concavity also assures the continuity of this function. Finally, minimization over finite $N$ hyperplanes guarantees that the function $g(\varpi)$ is made up of at most $N$ hyperplanes. 


\subsection{Proof of Lemma~\ref{lem:j_function}}

First, let us consider the function $\bar{J}_{K-1}(\varpi)$ given by: 
\begin{align}
\label{eq:convexity1}
\bar{J}_{K-1}(\varpi) =& \min \bigg [ g(\varpi), c_{K} + \sum_{F_{K}}  \Delta_{K}^T(F_{K}) \varpi  \nonumber\\
&\times  \bar{J}_{K} \bigg (  \frac{\diag \big(\Delta_{K}(F_{K}) \big) \varpi }{\Delta_{K}^T(F_{K}) \varpi} \bigg )  \bigg ]. 
\end{align}
Using the fact that $\bar{J}_{K}(\pi_{K}) = g(\pi_{K})$, we can rewrite Eq.~(\ref{eq:convexity1}) as follows: 
\begin{align}
\label{eq:convexity2}
\bar{J}_{K-1}(\varpi) =& \min \bigg [ g(\varpi), c_{K} + \sum_{F_{K}}  \Delta_{K}^T(F_{K}) \varpi  \nonumber \\
&\times g  \bigg(  \frac{\diag \big(\Delta_{K}(F_{K}) \big) \varpi }{\Delta_{K}^T(F_{K}) \varpi}  \bigg)  \bigg ]. 
\end{align}

We focus our attention on the following function inside the summation of Eq.~(\ref{eq:convexity2}):
\begin{align}
\label{eq:convexity6}
Q(\varpi) &\triangleq \Delta_{K}^T(F_{K}) \varpi 
g  \bigg( \frac{\diag \big(\Delta_{K}(F_{K}) \big) \varpi }{\Delta_{K}^T(F_{K}) \varpi} \bigg).
\end{align}
Using the definition of $g(\varpi)$ in  Eq.~(\ref{eq:g_function}), we can rewrite Eq.~(\ref{eq:convexity6}) as follows:
\begin{align}
Q(\varpi) &= \Delta_{K}^T(F_{K}) \varpi  
\min_{1 \leqslant j \leqslant N} \bigg [   M_j^T  \frac{\diag \big(\Delta_{K}(F_{K}) \big) \varpi }{\Delta_{K}^T(F_{K}) \varpi} \bigg ]
\nonumber \\
&= \min_{1 \leqslant j \leqslant N} \bigg [\Delta_{K}^T(F_{K}) \varpi M_j^T  \frac{\diag \big(\Delta_{K}(F_{K}) \big) \varpi }{\Delta_{K}^T(F_{K}) \varpi} \bigg ]  \nonumber \\
&= \min_{1 \leqslant j \leqslant N} \bigg [ M_j^T \diag \big(\Delta_{K}(F_{K}) \big) \varpi \bigg ].
\end{align}
Note that the term $M_j^T \diag \big(\Delta_{K}(F_{K}) \big) \varpi$ is linear with respect to $\varpi$. Using the fact that the minimum of linear functions is a concave, piecewise linear function, implies that $Q(\varpi)$ is a concave, piecewise linear function.  Furthermore, we recall: i) the non--negative sum of concave/piecewise linear functions is also a concave/piecewise linear function, and ii) the minimum of two concave/piecewise linear functions is also a concave/piecewise linear function. Based on these two facts, and the fact that $\Delta_{K}(F_{K})$ is a probability vector which is non--negative, we conclude that the function $\bar{J}_{K-1}(\varpi)$ in Eq.~(\ref{eq:convexity1}) is concave and piecewise linear. Concavity also assures the continuity of this function.

Then, let us consider the function $\bar{J}_{K-2}(\varpi)$ given by: 
\begin{align}
\label{eq:convexity7}
\bar{J}_{K-2}(\varpi) = & \min \bigg [ g(\varpi), c_{K-1} + \sum_{F_{K-1}}  \Delta_{K-1}^T(F_{K-1}) \varpi  \nonumber \\ &\times \bar{J}_{K-1} \bigg (  \frac{\diag \big(\Delta_{K-1}(F_{K-1}) \big) \varpi }{\Delta_{K-1}^T(F_{K-1}) \varpi} \bigg )  \bigg ]. 
\end{align}
We have already proved that the functions $\bar{J}_{K-1}(\varpi)$ and $g(\varpi)$ are concave and piecewise linear. Using the facts that i) the non--negative sum of concave/piecewise linear functions is also a concave/piecewise linear function, and ii) the minimum of two concave/piecewise linear functions is also a concave/piecewise linear function, we conclude that the function $\bar{J}_{K-2}(\varpi)$ is also concave and piecewise linear. Concavity also assures the continuity of this function. Using similar arguments, the concavity, the continuity and the piecewise linearity of functions $\bar{J}_{k}(\varpi), n=0,\dots,K-3$, can also be guaranteed.  

\subsection{Proof of Theorem~\ref{thm:alternative}}

Let us start the proof by showing that all $N$ corners of the $N-1$ dimensional unit simplex always correspond to stopping irrespective of the stage. In other words, when $\varpi = e_i$, $g(\varpi) < \bar{\mathcal{C}}_{k+1}(\varpi)$, 
for all $k=0,\dots,K-1$, where $e_i$ denotes the column vector with a $1$ in the  $i$th coordinate and $0$'s elsewhere. At stage $k = K-1$, we have that:
\begin{align}
\bar{\mathcal{C}}_{K}(\varpi) &= c_{K} + \sum_{F_{K}} \Delta_{K}^T(F_{K}) e_i  
\bar{J}_{K} \bigg (  \frac{\diag \big(\Delta_{K}(F_{K})  \big) e_i }{\Delta_{K}^T(F_{K}) e_i} \bigg ) \nonumber \\ &= c_{K} + \sum_{F_{K}} \Delta_{K}^T(F_{K}) e_i  
g \bigg (  \frac{\diag \big(\Delta_{K}(F_{K})  \big) e_i }{\Delta_{K}^T(F_{K}) e_i} \bigg ) \nonumber \\
&= c_{K} + \sum_{F_{K}} P(F_{K}| T_i) 
g \bigg (  \frac{ P(F_{K}| T_i) e_i }{P(F_{K}| T_i)} \bigg ) \nonumber \\
&= c_{K} +  g (  e_i  ) \sum_{F_{K}} P(F_{K}| T_i) \nonumber \\
&= c_{K} +  g (  e_i  ) \nonumber \\
&> g (  e_i  ),
\end{align}
where the last inequality holds since $c_K >0$. From Eq.~(\ref{eq:convexity1}), we see that $\bar{J}_{K-1}(e_i)  = g (  e_i  )$. Then, let us consider the case $k = K-2$ as follows:
\begin{align}
\bar{\mathcal{C}}_{K-1}(\varpi) &= c_{K-1} + \sum_{F_{K-1}} \Delta_{K-1}^T(F_{K-1}) e_i  \nonumber \\
&\times  \bar{J}_{K-1} \bigg (  \frac{\diag \big(\Delta_{K-1}(F_{K-1})  \big) e_i }{\Delta_{K-1}^T(F_{K-1}) e_i} \bigg ) \nonumber \\
&= c_{K-1} + \sum_{F_{K-1}} P(F_{K-1}| T_i)  
\bar{J}_{K-1} \bigg (  \frac{ P(F_{K-1}| T_i) e_i }{P(F_{K-1}| T_i)} \bigg ) \nonumber \\
&= c_{K-1} +  
\bar{J}_{K-1} ( e_i  ) \sum_{F_{K-1}} P(F_{K-1}| T_i) \nonumber \\
&= c_{K-1} +  g (  e_i  ) \sum_{F_{K-1}} P(F_{K-1}| T_i) \nonumber \\
&= c_{K-1} +  g (  e_i  ) \nonumber \\
&> g (  e_i  ),
\end{align}
where the last inequality holds since $c_{K-1} >0$. Using similar arguments, the latter result can be proven for all $k=0,\dots,K-3$. 
The rest of the proof is very intuitive. Using the facts: (i) the functions  $\bar{\mathcal{C}}_{k+1}(\varpi)$ are concave (see the proof of Lemma~\ref{lem:j_function}), and (ii) the simplex  corners always correspond to stopping, we see that  the
hyperplanes of $g(\varpi)$ connected to the $N$ corners of the unit simplex can have only one intersection with each $\bar{\mathcal{C}}_{k+1}(\varpi)$. Finally, using the fact that $g(\varpi)$ is made up of at most $N$ hyperplanes (see Lemma~\ref{lem:g_function}), we conclude that at every stage $k$, there are at most $N$ threshold curves which split up the probability space of $\varpi$ (i.e., the $N-1$ dimensional unit simplex)  into areas that correspond to either continuing or stopping. 


\ifCLASSOPTIONcompsoc
\section*{Acknowledgments}
\else
\section*{Acknowledgment}
\fi
This material is based upon work supported by the National Science
Foundation under Grant No. ECCS–-1737443.

\ifCLASSOPTIONcaptionsoff
\newpage
\fi

\bibliographystyle{IEEEtran}
\begin{small}
	\bibliography{IEEEabrv,references}

\begin{thebibliography}{10}
\providecommand{\url}[1]{#1}
\csname url@samestyle\endcsname
\providecommand{\newblock}{\relax}
\providecommand{\bibinfo}[2]{#2}
\providecommand{\BIBentrySTDinterwordspacing}{\spaceskip=0pt\relax}
\providecommand{\BIBentryALTinterwordstretchfactor}{4}
\providecommand{\BIBentryALTinterwordspacing}{\spaceskip=\fontdimen2\font plus
\BIBentryALTinterwordstretchfactor\fontdimen3\font minus
  \fontdimen4\font\relax}
\providecommand{\BIBforeignlanguage}[2]{{%
\expandafter\ifx\csname l@#1\endcsname\relax
\typeout{** WARNING: IEEEtran.bst: No hyphenation pattern has been}%
\typeout{** loaded for the language `#1'. Using the pattern for}%
\typeout{** the default language instead.}%
\else
\language=\csname l@#1\endcsname
\fi
#2}}
\providecommand{\BIBdecl}{\relax}
\BIBdecl

\bibitem{guyon2003introduction}
I.~Guyon and A.~Elisseeff, ``An introduction to variable and feature
  selection,'' \emph{Journal of machine learning research}, vol.~3, no. Mar,
  pp. 1157--1182, 2003.

\bibitem{liu2005toward}
H.~Liu and L.~Yu, ``Toward integrating feature selection algorithms for
  classification and clustering,'' \emph{IEEE Transactions on Knowledge \& Data
  Engineering}, no.~4, pp. 491--502, 2005.

\bibitem{perkins2003online}
S.~Perkins and J.~Theiler, ``Online feature selection using grafting,'' in
  \emph{Proceedings of the 20th International Conference on Machine Learning
  (ICML-03)}, 2003, pp. 592--599.

\bibitem{zhou2005streaming}
J.~Zhou, D.~Foster, R.~Stine, and L.~Ungar, ``Streaming feature selection using
  alpha-investing,'' in \emph{Proceedings of the eleventh ACM SIGKDD
  international conference on Knowledge discovery in data mining}.\hskip 1em
  plus 0.5em minus 0.4em\relax ACM, 2005, pp. 384--393.

\bibitem{wu2012online}
X.~Wu, K.~Yu, W.~Ding, H.~Wang, and X.~Zhu, ``Online feature selection with
  streaming features,'' \emph{IEEE transactions on pattern analysis and machine
  intelligence}, vol.~35, no.~5, pp. 1178--1192, 2012.

\bibitem{yu2014towards}
K.~Yu, X.~Wu, W.~Ding, and J.~Pei, ``Towards scalable and accurate online
  feature selection for big data,'' in \emph{2014 IEEE International Conference
  on Data Mining}.\hskip 1em plus 0.5em minus 0.4em\relax IEEE, 2014, pp.
  660--669.

\bibitem{zhou2019ofs}
P.~Zhou, X.~Hu, P.~Li, and X.~Wu, ``Ofs-density: A novel online streaming
  feature selection method,'' \emph{Pattern Recognition}, vol.~86, pp. 48--61,
  2019.

\bibitem{eskandari2016online}
S.~Eskandari and M.~M. Javidi, ``Online streaming feature selection using rough
  sets,'' \emph{International Journal of Approximate Reasoning}, vol.~69, pp.
  35--57, 2016.

\bibitem{zhou2019online}
P.~Zhou, X.~Hu, P.~Li, and X.~Wu, ``Online streaming feature selection using
  adapted neighborhood rough set,'' \emph{Information Sciences}, vol. 481, pp.
  258--279, 2019.

\bibitem{javidi2019online}
M.~M. Javidi and S.~Eskandari, ``Online streaming feature selection: a minimum
  redundancy, maximum significance approach,'' \emph{Pattern Analysis and
  Applications}, vol.~22, no.~3, pp. 949--963, 2019.

\bibitem{hoi2012online}
S.~C. Hoi, J.~Wang, P.~Zhao, and R.~Jin, ``Online feature selection for mining
  big data,'' in \emph{Proceedings of the 1st international workshop on big
  data, streams and heterogeneous source mining: Algorithms, systems,
  programming models and applications}.\hskip 1em plus 0.5em minus 0.4em\relax
  ACM, 2012, pp. 93--100.

\bibitem{wang2014onine}
J.~{Wang}, P.~{Zhao}, S.~C.~H. {Hoi}, and R.~{Jin}, ``Online feature selection
  and its applications,'' \emph{IEEE Transactions on Knowledge and Data
  Engineering}, vol.~26, no.~3, pp. 698--710, March 2014.

\bibitem{wu2017large}
Y.~Wu, S.~C. Hoi, T.~Mei, and N.~Yu, ``Large-scale online feature selection for
  ultra-high dimensional sparse data,'' \emph{ACM Transactions on Knowledge
  Discovery from Data (TKDD)}, vol.~11, no.~4, p.~48, 2017.

\bibitem{li2018feature}
J.~Li, K.~Cheng, S.~Wang, F.~Morstatter, R.~P. Trevino, J.~Tang, and H.~Liu,
  ``Feature selection: A data perspective,'' \emph{ACM Computing Surveys
  (CSUR)}, vol.~50, no.~6, p.~94, 2018.

\bibitem{wang2017feature}
J.~Wang, J.-M. Wei, Z.~Yang, and S.-Q. Wang, ``Feature selection by maximizing
  independent classification information,'' \emph{IEEE Transactions on
  Knowledge and Data Engineering}, vol.~29, no.~4, pp. 828--841, 2017.

\bibitem{zhang2018muse}
Z.~Zhang and K.~K. Parhi, ``Muse: Minimum uncertainty and sample elimination
  based binary feature selection,'' \emph{IEEE Transactions on Knowledge and
  Data Engineering}, 2018.

\bibitem{chandrashekar2014survey}
G.~Chandrashekar and F.~Sahin, ``A survey on feature selection methods,''
  \emph{Computers \& Electrical Engineering}, vol.~40, no.~1, pp. 16--28, 2014.

\bibitem{saeys2007review}
Y.~Saeys, I.~Inza, and P.~Larra{\~n}aga, ``A review of feature selection
  techniques in bioinformatics,'' \emph{bioinformatics}, vol.~23, no.~19, pp.
  2507--2517, 2007.

\bibitem{hu2018survey}
X.~Hu, P.~Zhou, P.~Li, J.~Wang, and X.~Wu, ``A survey on online feature
  selection with streaming features,'' \emph{Frontiers of Computer Science},
  vol.~12, no.~3, pp. 479--493, 2018.

\bibitem{alnuaimi2019streaming}
N.~AlNuaimi, M.~M. Masud, M.~A. Serhani, and N.~Zaki, ``Streaming feature
  selection algorithms for big data: A survey,'' \emph{Applied Computing and
  Informatics}, 2019.

\bibitem{bolon2014framework}
V.~Bol{\'o}n-Canedo, I.~Porto-D{\'\i}az, N.~S{\'a}nchez-Maro{\~n}o, and
  A.~Alonso-Betanzos, ``A framework for cost-based feature selection,''
  \emph{Pattern Recognition}, vol.~47, no.~7, pp. 2481--2489, 2014.

\bibitem{shu2016multi}
W.~Shu and H.~Shen, ``Multi-criteria feature selection on cost-sensitive data
  with missing values,'' \emph{Pattern Recognition}, vol.~51, pp. 268--280,
  2016.

\bibitem{rosenblatt1958perceptron}
F.~Rosenblatt, ``The perceptron: a probabilistic model for information storage
  and organization in the brain.'' \emph{Psychological review}, vol.~65, no.~6,
  p. 386, 1958.

\bibitem{novikoff1963convergence}
A.~B. Novikoff, ``On convergence proofs for perceptrons,'' STANFORD RESEARCH
  INST MENLO PARK CA, Tech. Rep., 1963.

\bibitem{crammer2006online}
K.~Crammer, O.~Dekel, J.~Keshet, S.~Shalev-Shwartz, and Y.~Singer, ``Online
  passive-aggressive algorithms,'' \emph{Journal of Machine Learning Research},
  vol.~7, no. Mar, pp. 551--585, 2006.

\bibitem{duchi2011adaptive}
J.~Duchi, E.~Hazan, and Y.~Singer, ``Adaptive subgradient methods for online
  learning and stochastic optimization,'' \emph{Journal of Machine Learning
  Research}, vol.~12, no. Jul, pp. 2121--2159, 2011.

\bibitem{van2016metagrad}
T.~van Erven and W.~M. Koolen, ``Metagrad: Multiple learning rates in online
  learning,'' in \emph{Advances in Neural Information Processing Systems},
  2016, pp. 3666--3674.

\bibitem{luo2016efficient}
H.~Luo, A.~Agarwal, N.~Cesa-Bianchi, and J.~Langford, ``Efficient second order
  online learning by sketching,'' in \emph{Advances in Neural Information
  Processing Systems}, 2016, pp. 902--910.

\bibitem{zhang2018adaptive}
L.~Zhang, S.~Lu, and Z.-H. Zhou, ``Adaptive online learning in dynamic
  environments,'' in \emph{Advances in Neural Information Processing Systems},
  2018, pp. 1323--1333.

\bibitem{li2002relaxed}
Y.~Li and P.~M. Long, ``The relaxed online maximum margin algorithm,''
  \emph{Machine Learning}, vol.~46, no.~1, pp. 361--387, 2002.

\bibitem{wang2013cost}
J.~Wang, P.~Zhao, and S.~C. Hoi, ``Cost-sensitive online classification,''
  \emph{IEEE Transactions on Knowledge and Data Engineering}, vol.~26, no.~10,
  pp. 2425--2438, 2013.

\bibitem{zhao2018adaptive}
P.~Zhao, Y.~Zhang, M.~Wu, S.~C. Hoi, M.~Tan, and J.~Huang, ``Adaptive
  cost-sensitive online classification,'' \emph{IEEE Transactions on Knowledge
  and Data Engineering}, vol.~31, no.~2, pp. 214--228, 2018.

\bibitem{hoi2018online}
S.~C. Hoi, D.~Sahoo, J.~Lu, and P.~Zhao, ``Online learning: A comprehensive
  survey,'' \emph{arXiv preprint arXiv:1802.02871}, 2018.

\bibitem{zhang2016short}
C.~Zhang, H.~Wei, J.~Zhao, T.~Liu, T.~Zhu, and K.~Zhang, ``Short-term wind
  speed forecasting using empirical mode decomposition and feature selection,''
  \emph{Renewable Energy}, vol.~96, pp. 727--737, 2016.

\bibitem{yang2013feature}
S.~Yang, ``On feature selection for traffic congestion prediction,''
  \emph{Transportation Research Part C: Emerging Technologies}, vol.~26, pp.
  160--169, 2013.

\bibitem{lee2009using}
M.-C. Lee, ``Using support vector machine with a hybrid feature selection
  method to the stock trend prediction,'' \emph{Expert Systems with
  Applications}, vol.~36, no.~8, pp. 10\,896--10\,904, 2009.

\bibitem{ding2005minimum}
C.~Ding and H.~Peng, ``Minimum redundancy feature selection from microarray
  gene expression data,'' \emph{Journal of bioinformatics and computational
  biology}, vol.~3, no.~02, pp. 185--205, 2005.

\bibitem{seo2014feature}
J.-H. Seo, Y.~H. Lee, and Y.-H. Kim, ``Feature selection for very short-term
  heavy rainfall prediction using evolutionary computation,'' \emph{Advances in
  Meteorology}, vol. 2014, 2014.

\bibitem{shiryaev2007optimal}
A.~N. Shiryaev, \emph{{Optimal Stopping Rules}}.\hskip 1em plus 0.5em minus
  0.4em\relax Springer Science \& Business Media, 2007, vol.~8.

\bibitem{BertsekasDPOC05}
D.~P. Bertsekas, \emph{{Dynamic Programming and Optimal Control}}.\hskip 1em
  plus 0.5em minus 0.4em\relax Athena Scientific, 2005, vol.~1.

\bibitem{yang2006stable}
K.~Yang, Z.~Cai, J.~Li, and G.~Lin, ``A stable gene selection in microarray
  data analysis,'' \emph{BMC bioinformatics}, vol.~7, no.~1, p. 228, 2006.

\bibitem{spall2005introduction}
J.~C. Spall, \emph{Introduction to stochastic search and optimization:
  estimation, simulation, and control}.\hskip 1em plus 0.5em minus 0.4em\relax
  John Wiley \& Sons, 2005, vol.~65.

\bibitem{pflug2012optimization}
G.~C. Pflug, \emph{Optimization of stochastic models: the interface between
  simulation and optimization}.\hskip 1em plus 0.5em minus 0.4em\relax Springer
  Science \& Business Media, 2012, vol. 373.

\bibitem{NIPSdataset}
``{Colpinet: Feature Selection Challenge},'' [Online]. Available:
  \url{http://clopinet.com/isabelle/Projects/NIPS2003/}.

\bibitem{Libsvm}
``{LIBSVM data sets},'' [Online]. Available:
  \url{https://www.csie.ntu.edu.tw/~cjlin/libsvmtools/datasets/}.

\end{thebibliography}
\end{small}

%
\newpage
\begin{IEEEbiography}[{\includegraphics[width=1.16in,height=1.16in,clip,keepaspectratio]{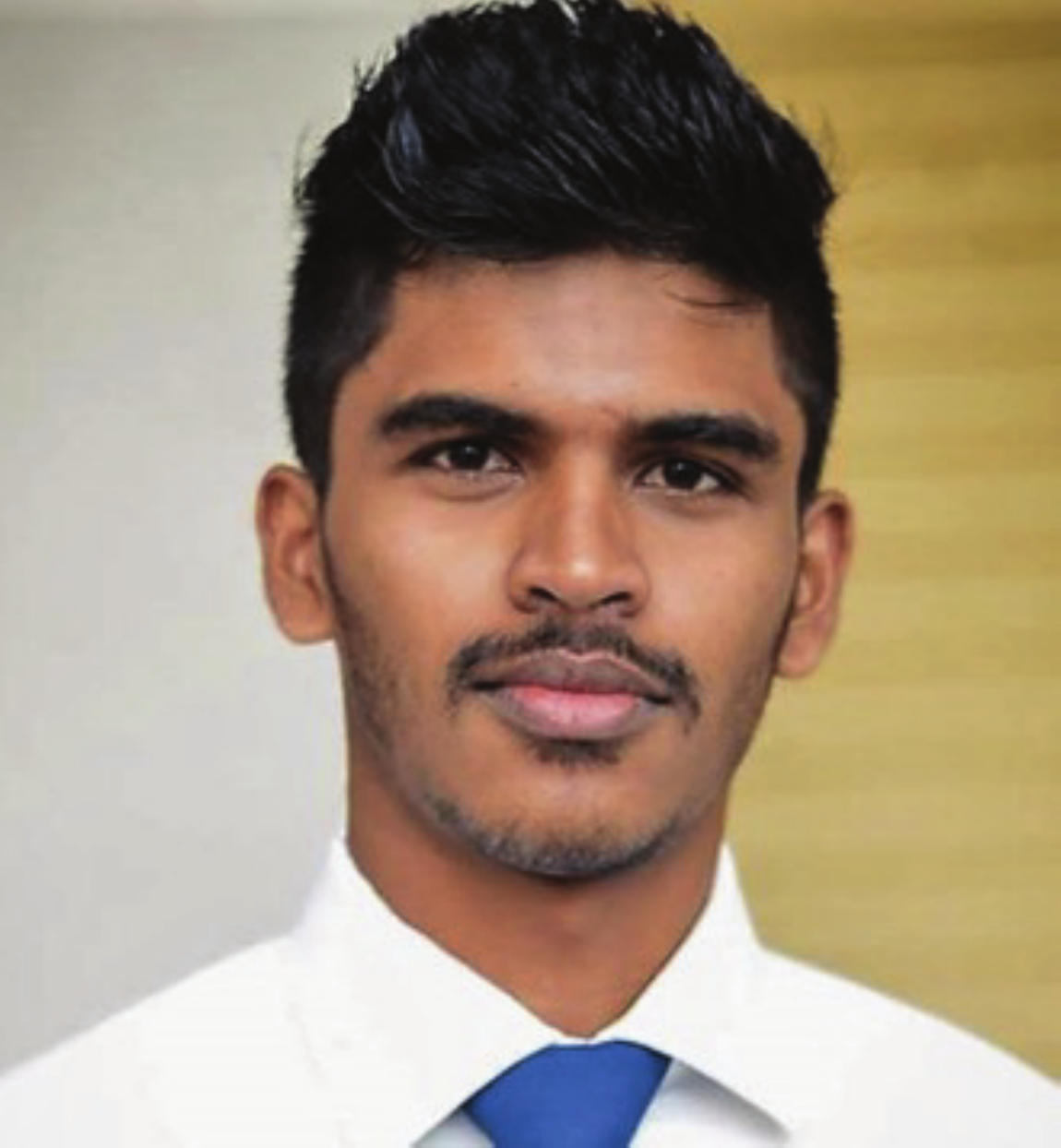}}]
	{Yasitha Warahena Liyanage} received the B.S. degree in electrical and electronic engineering from the University of Peradeniya, Sri Lanka, in 2016. Currently, he is working toward the Ph.D. degree in electrical and computer engineering at the University at Albany, SUNY. His research interests include quickest change detection, optimal stopping theory and machine learning.
\end{IEEEbiography}
\vskip -2\baselineskip plus -1fil
\begin{IEEEbiography}[{\includegraphics[width=1.25in,height=1.25in,clip,keepaspectratio]{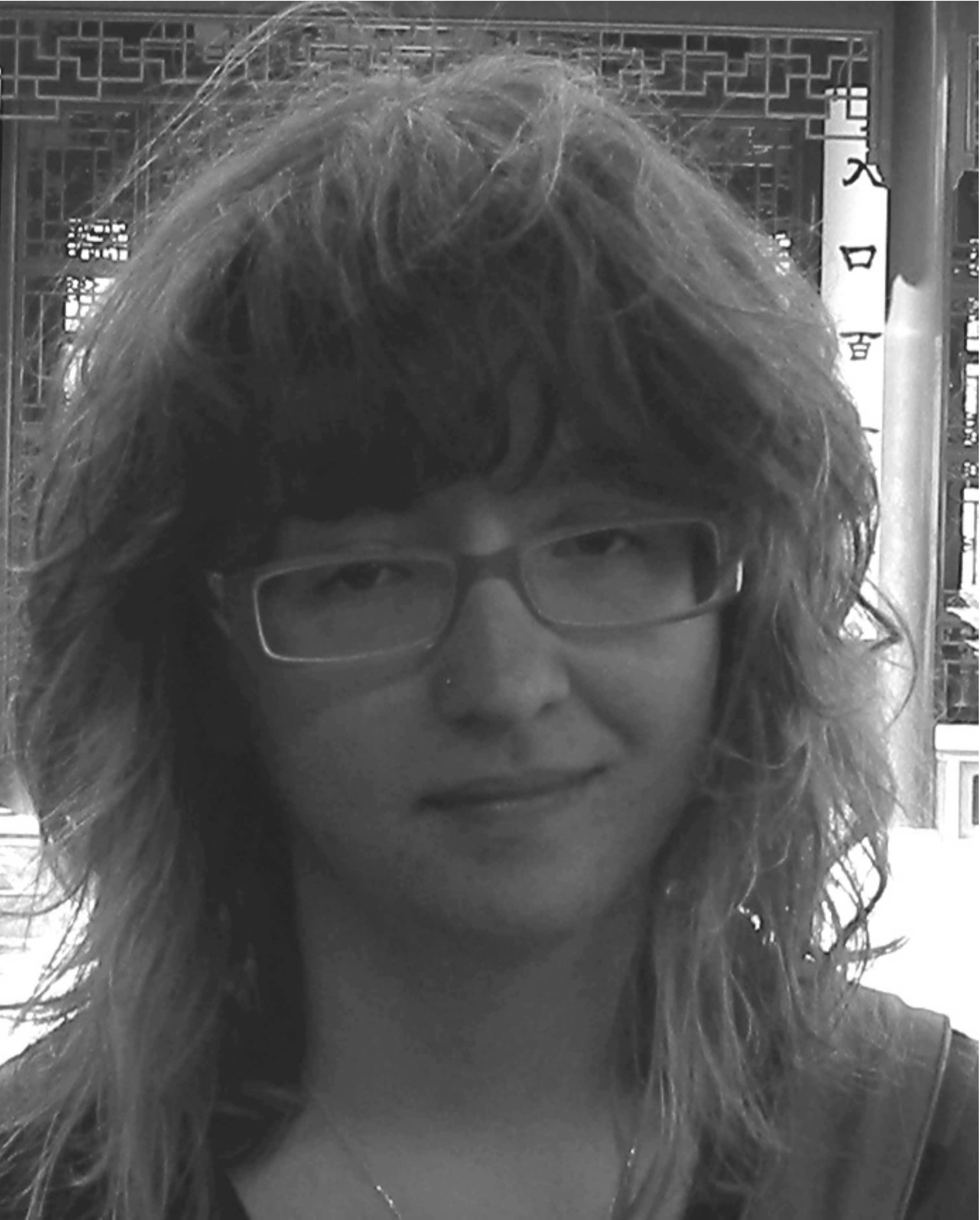}}]
	{Daphney-Stavroula Zois} received the B.S. degree in computer engineering and informatics from the University of Patras, Patras, Greece, and the M.S. and Ph.D. degrees in electrical engineering from the University of Southern California, Los Angeles, CA, USA. Previous appointments include the University of Illinois, Urbana--Champaign, IL, USA. She is an Assistant Professor in the Department of Electrical and Computer Engineering, University at Albany, State University of New York, Albany, NY, USA. She received the Viterbi Dean's and Myronis Graduate Fellowships. She has served and is serving as Co-Chair, TPC member or reviewer in international conferences and journals, such as GlobalSIP, Globecom, and ICASSP, and IEEE Transactions on Signal Processing. Her research interests include decision making under uncertainty, machine learning, detection \& estimation theory, intelligent systems design, and signal processing.
\end{IEEEbiography}
\vskip -2\baselineskip plus -1fil
\begin{IEEEbiography}[{\includegraphics[width=1.2in,height=1.2in,clip,keepaspectratio]{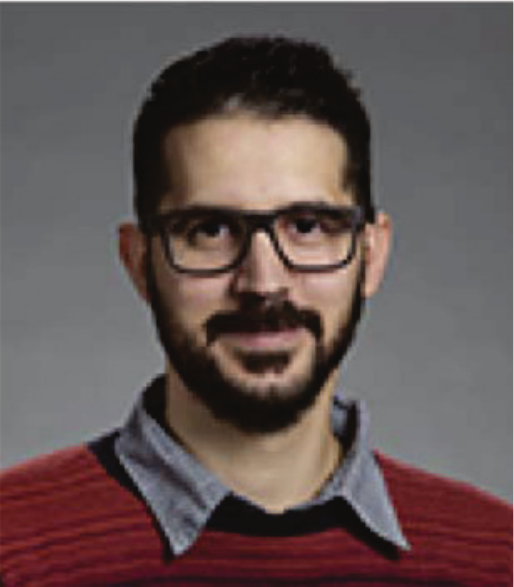}}]
	{Charalampos Chelmis} is an Assistant Professor in Computer Science at the University at Albany, State University of New York, and the director of the Intelligent Big Data Analytics, Applications, and Systems (IDIAS) Lab, focusing on problems involving big, often networked, data. He has served and is serving as Co-Chair, TPC member or reviewer in numerous international conferences and journals such as ASONAM, SocInfo, and ICWSM. Currently, he serves as Associate Editor of the Journal of Parallel and Distributed Systems, and served as Guest Editor for the Encyclopedia of Social Network Analysis and Mining. He received the B.S. degree in computer engineering and informatics from the University of Patras, Greece in 2007, and the M.S. and Ph.D. degrees in computer science from the University of Southern California in 2010 and 2013, respectively.
\end{IEEEbiography}



\end{document}